\pdfoutput=1

\documentclass[11pt]{article}

\usepackage{acl}

\usepackage{times}
\usepackage{latexsym}

\usepackage[T1]{fontenc}

\usepackage[utf8]{inputenc}

\usepackage{microtype}

\usepackage{inconsolata}

\usepackage{graphicx}

\usepackage{amsmath} 
\usepackage{amssymb}  
\usepackage{booktabs}    
\usepackage{multirow}    
\usepackage{array}       
\usepackage{makecell}    
\usepackage{adjustbox}   

\usepackage{caption}
\usepackage{enumitem}

\usepackage{algorithm}
\usepackage{algpseudocode}

\title{Domain$o1$s: Guiding LLM Reasoning for Explainable Answers \\in High-Stakes Domains}

\author{
  \textbf{Xu Chu\textsuperscript{†}},
  \textbf{Zhijie Tan\textsuperscript{†}},
  \textbf{Hanlin Xue},
  \textbf{Guanyu Wang},
  \textbf{Tong Mo},
  \textbf{Weiping Li\textsuperscript{*}}
\\
\\
  \textsuperscript{1}School of Software and Microelectronics, Peking University, Beijing, China
\\
  \href{mailto:chuxu@stu.pku.edu.cn}{chuxu@stu.pku.edu.cn},
  \href{mailto:besttangent@stu.pku.edu.cn}{besttangent@stu.pku.edu.cn},
  \href{mailto:colinneverland@stu.pku.edu.cn}{colinneverland@stu.pku.edu.cn}
\\
  \href{mailto:2301210418@stu.pku.edu.cn}{2301210418@stu.pku.edu.cn},
  \href{mailto:motong@ss.pku.edu.cn}{motong@ss.pku.edu.cn},
  \href{mailto:wpli@ss.pku.edu.cn}{wpli@ss.pku.edu.cn}
\\
  \small{
    \textsuperscript{†}Equal contribution
    \hspace{2em}
    \textsuperscript{*}Corresponding author
  }
}

\begin{document}
\maketitle
\begin{abstract}
Large Language Models (LLMs) are widely applied to downstream domains. However, current LLMs for high-stakes domain tasks, such as financial investment and legal QA, typically generate brief answers without reasoning processes and explanations. This limits users' confidence in making decisions based on their responses. While original CoT shows promise, it lacks self-correction mechanisms during reasoning. This work introduces Domain$o1$s, which enhances LLMs' reasoning capabilities on domain tasks through supervised fine-tuning and tree search. We construct CoT-stock-2k and CoT-legal-2k datasets for fine-tuning models that activate domain-specific reasoning steps based on their judgment. 
Additionally, we propose Selective Tree Exploration to spontaneously explore solution spaces and sample optimal reasoning paths to improve performance. We also introduce PROOF-Score, a new metric for evaluating domain models' explainability, complementing traditional accuracy metrics with richer assessment dimensions. Extensive experiments on stock investment recommendation and legal reasoning QA tasks demonstrate Domain$o1$s's leading performance and explainability. Our code is available at \url{https://github.com/Hyalinesky/Domaino1s}.
\end{abstract}

\section{Introduction}
\begin{figure}[t]
\centering
\includegraphics[width=0.49\textwidth]{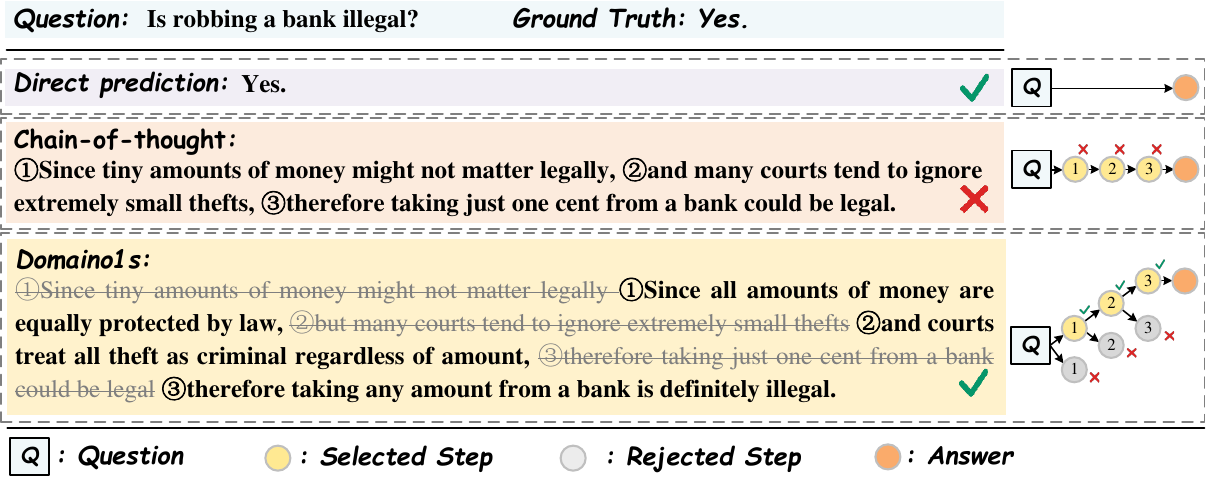} 
\caption{Comparison of Domain$o1$s and other paradigms on a demonstrative example. Domain$o1$s expands reasoning paths and obtains optimal ones through tree search.}
\label{intro_v2}
\end{figure}

In specific domains such as finance~\cite{xing2024designing,jeong2024fine,cheng2024adapting}, law~\cite{cheong2024not,colombo2024saullm}, and biomedicine~\cite{labrak-etal-2024-biomistral,wang2023huatuo}, Large Language Models (LLMs) are widely used for tasks like recommendation (e.g., stock investment recommendation~\cite{koa2024learning,qin2024followakoinvestor,takayanagi2023personalized}) and question answering (e.g., legal reasoning QA~\cite{guha2024legalbench,wang-etal-2023-maud,ujwal2024reasoning}). However, popular approaches mainly adopt direct prediction paradigms that immediately generate brief answers to questions~\cite{Cheng-etal-2024-instruction,cheng2024adapting,yue2023disc}, leading to answers lacking explainability. In practical applications within high-stakes domains like finance and law, users may not trust results lacking explainability~\cite{biranhuman2017} to guide decision-making. While Chain-of-Thought (CoT) reasoning demonstrates the ability to enhance models' step-by-step thinking and domain problem solving~\cite{li-etal-2024-alphafin,jiang2023legal,miao2024chain} and provides explainable reasoning processes, its single-pass generated reasoning chains lack error correction mechanisms. If errors occur in early reasoning steps, the model continues reasoning along the flawed path, affecting the subsequent reasoning process, as shown in Figure~\ref{intro_v2}. This poses challenges for solving domain tasks, as flawed reasoning processes may introduce legal and ethical risks.

Recently introduced o1-type models~\cite{openai2024learning,openO1team2024,zhao2024marco}, with their exceptional reasoning capabilities, demonstrate powerful performance surpassing reasoning methods like CoT in mathematics, physics, and coding. Compared to LLMs using CoT, o1-type models feature longer reasoning chains and reasoning time. They are considered to perform multi-stage reasoning rather than generating complete reasoning chains in single-pass, which enhances the accuracy of LLM reasoning. However, despite high-stakes domains requiring high-quality reasoning, extending o1-type models' capabilities to these domains remains an unexplored research gap.

In this paper, we design Domain$o1$s to provide explainable answers for high-stakes domain problems. Domain$o1$s includes two model variants, Domain$o1$s-finance and Domain$o1$s-legal. As shown in Figure~\ref{intro_v2}, Domain$o1$s can (1) perform autonomous step-by-step reasoning, and (2) expand reasoning paths through tree search to obtain optimal ones. To achieve (1), we use GPT-4o~\cite{hurst2024gpt} to generate CoT data and construct CoT-stock-2k and CoT-legal-2k datasets for supervised fine-tuning. During dataset construction, we employ 26 special tokens (e.g., <SUMMARY>) to prompt GPT-4o to distinguish different steps in the reasoning process explicitly. In the supervised fine-tuning process, we remove these special tokens from the answers, enabling the model to autonomously select and organize intermediate steps in the reasoning chain. To achieve (2) during answer generation, we introduce a novel Selective Tree Exploration method to find the optimal reasoning paths. This method uses the average perplexity of tokens in each reasoning step to decide whether to explore new paths and select the best path. Compared to traditional search methods~\cite{weng2022large,jiang2023tigerscore, chen2024sentence}, Selective Tree Exploration balances search performance and time cost. We evaluate Domains$o1$s on stock investment recommendation~\cite{koa2024learning} and legal reasoning QA~\cite{guha2024legalbench} datasets. Unlike most domain benchmarks~\cite{koa2024learning,yang2022numhtml,guha2024legalbench}, we point out that focusing solely on answer accuracy makes it difficult to determine whether models properly reason through given contexts rather than relying on shortcuts or overfitting. We emphasize the necessity of evaluating domain models' explainability and introduce a new evaluation metric PROOF-Score (\underline{P}rincipled \underline{r}ating for reas\underline{o}ning completeness, d\underline{o}main safety, and \underline{f}actual accuracy) to fill this gap. Results show that Domain$o1$s improves reasoning accuracy while providing high-quality, explainable reasoning processes. Our contributions are:

• Domain$o1$s is proposed for explainable answers, with two model variants.

• CoT-stock-2k and CoT-legal-2k datasets are constructed for fine-tuning. Selective Tree Exploration is proposed as a reasoning path search method that balances performance and time cost.

• PROOF-Score is proposed to evaluate the explainability of domain model answers, introducing a new perspective for domain model evaluation.

• Domain$o1$s achieves leading performance, demonstrating the effectiveness of its reasoning capabilities in solving high-stakes domain tasks.

\begin{figure*}[t]
\centering
\includegraphics[width=0.99\textwidth]{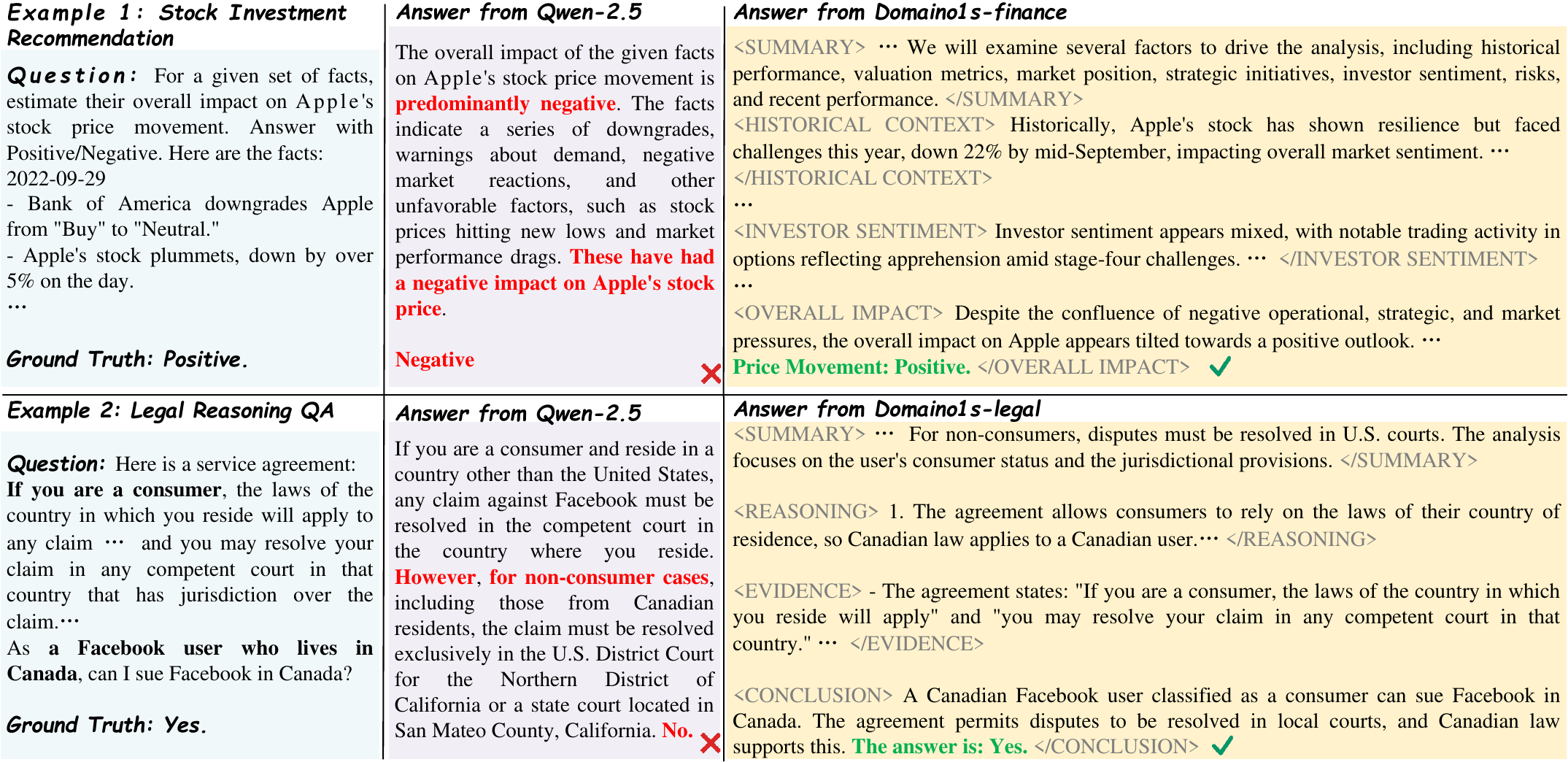} 
\caption{Comparison between the base model Qwen-2.5-Instruct~\cite{qwen2.5} and Domain$o1$s. The base model shows notable reasoning errors. In contrast, Domain$o1$s breaks problems into multiple reasoning steps and reaches well-supported conclusions through systematic analysis. Details in Appendix~\ref{Appendix_Answer_Demonstration}.}
\label{Domain-o1}
\end{figure*}

\section{Related Works}
\subsection{LLMs for Specific Domains}
LLM applications in specific domains typically follow three approaches: training from scratch, fine-tuning, and prompt learning. While training from scratch (e.g., BloombergGPT~\cite{wu2024bloomberggpt}) shows promising results, it requires significant computational resources and data~\cite{yang2023fingpt,zhao2023beyond,xie2023pixiu}. Fine-tuning emerges as a cost-effective alternative, with researchers using GPT-4~\cite{li2024llava} or low-cost automated methods~\cite{cheng2024adapting,koa2024learning} to generate fine-tuning data. Prompt learning methods enhance model capabilities without parameter modification through template engineering or knowledge retrieval~\cite{Li2023ChatDoctorAM,chatlaw,huang2023lawyer}, such as CoT~\cite{wei2022chain} reasoning. o1-type models are typically constructed to equip LLMs with CoT reasoning capabilities through fine-tuning, followed by multi-pass search to obtain better reasoning paths.

\subsection{Single-Pass vs. Multi-Pass}
Prompt-based methods like CoT~\cite{wei2022chain,zhang2022automatic,lyu2023faithful} improve single-pass reasoning through better prompt templates. However, errors in intermediate reasoning steps can propagate through the chain. In contrast, search-based methods explore multiple reasoning paths in the solution space, treating each reasoning step as a node in the tree, and selecting the optimal path to improve reasoning quality~\cite{qi2024mutual}.

\subsection{Sampling Reasoning Paths}
Research on mathematical reasoning~\cite{brown2024large,wang-etal-2024-math} indicates that sampling different reasoning paths can improve performance compared to greedy one-time decoding. Best-of-N search~\cite{weng2022large,jiang2023tigerscore} generates N complete answers, allowing LLM to select the best response based on final results, but may miss high-quality intermediate reasoning steps~\cite{xie2024monte,chen-etal-2024-step}. Sentence-level Beam Search~\cite{chen2024sentence} generates multiple candidate sentences, selects the best one, and iteratively continues this process, but may get stuck in local optima. Stage-level Beam Search~\cite{xu2024llava} offers a compromise by generating and selecting optimal candidates for each reasoning step rather than sentences. 

\section{Methodology}
In this section, we first present the formal definition of LLM-based multi-step reasoning. Then, we introduce Domain$o1$s from two aspects: enhancing reasoning capabilities and solution expansion \& sampling. For aspect 1, Domain$o1$s facilitates a progressive reasoning process. For aspect 2, Domain$o1$s improves reasoning performance through tree search to obtain optimal reasoning paths. A comparison of reasoning examples with the base model is shown in Figure~\ref{Domain-o1}.

\subsection{Preliminaries}
For a given question $q$, the solution process can be decomposed into multiple reasoning steps. Consider a complete solution consisting of up to $T$ reasoning steps. The state $S_t$ comprising all reasoning steps from step 0 to $t$ can be represented as:
\vspace{-8pt}
\begin{equation}
S_t = \{s_0, s_1, \dots, s_t\}, 0 \leq t < T, t \in \mathbb{Z},
\end{equation}
where $s_t$ represents the $t$-th reasoning step, state $S_t$ represents the collection of reasoning processes from step 0 to $t$. An action $a_t (0 \leq t < T-1)$ is defined as choosing the next reasoning step $s_{t+1}$. The LLM constitutes a policy model, where the transition $f(S_{t+1}|a_t,S_t)$ from one state to the next is implemented by auto-regressively generating $s_{t+1}$ through the input sequence. To guide the LLM in selecting more reasonable subsequent reasoning step $s_{t+1}$, a value function $V(s_{t+1})$ is defined to evaluate the expected return of LLM's strategy.

\subsection{Enhancing Reasoning Capabilities}
\label{sectionEnhancing Reasoning Capabilities}
To enhance Domain$o1$s's reasoning capabilities in high-stakes domains (finance and legal), we employ supervised fine-tuning to let the model generate CoT-style responses. Since existing domain datasets or databases lack the detailed reasoning processes required for training Domain$o1$s models, we constructed two new datasets, CoT-stock-2k and CoT-legal-2k, using the training sets from stock investment recommendation~\cite{koa2024learning} and legal reasoning QA~\cite{guha2024legalbench,li2022parameter} datasets respectively. The construction details are as follows:

\textbf{Stock Investment Recommendation.} Contains price data and tweet information from the top 5 stocks across 11 industries during 2020-2022. The task is to predict stock price movement (positive or negative) for the next trading day based on facts extracted from tweets over the past 5 days. Due to the high volume of daily tweets, we fine-tuned Qwen-2.5-Instruct~\cite{qwen2.5} to generate daily tweet summaries. We utilized GPT-4o~\cite{hurst2024gpt} to generate CoT data, explicitly prompting it to decompose the answer generation process into 10 structured reasoning steps, including market factors~\cite{fama1993common}, company strategies~\cite{porter1985advantage}, and investor sentiment~\cite{baker2006investor}:

• \textbf{Summary}: Extract key facts from tweets about question $q$, identify main analysis focus.
• \textbf{Historical context}: Review historical performance and market context.
• \textbf{Valuation}: Assess current valuation metrics (e.g., P/E, price targets, market views).
• \textbf{Market size and dominance}: Evaluate company's industry standing and influence.
• \textbf{Strategic initiatives}: Review recent strategic moves (partnerships, innovation) and growth potential.
• \textbf{Investor sentiment}: Gauge investor mood through trading patterns and market discussion.
• \textbf{Risks and concerns}: Identify key investor concerns and risk factors.
• \textbf{Recent performance}: Analyze recent price movements and drivers.
• \textbf{Consolidation}: Review financial/stock structure changes (buybacks, profitability).
• \textbf{Overall impact}: Synthesize all analysis points, clearly indicate overall impact, and provide a final prediction (positive or negative) for the next trading day's stock price.

\textbf{Legal Reasoning QA}. Includes legal reasoning questions across multiple categories such as legal rule application, reasoning, and legal question classification, presented as multiple choice or true/false questions. We utilized GPT-4o~\cite{hurst2024gpt} to generate CoT data, explicitly prompting it to decompose the answer generation process into 4 structured reasoning steps:

• \textbf{Summary}: Extract key points from question $q$ and identify analysis focus.
• \textbf{Reasoning}: Apply step-by-step logic to reach answers. 
• \textbf{Evidence}: Systematically present supporting text and verify reasoning.
• \textbf{Conclusion}: Synthesize the analysis and state the final answer.

\begin{figure}[t]
\centering
\includegraphics[width=0.49\textwidth]{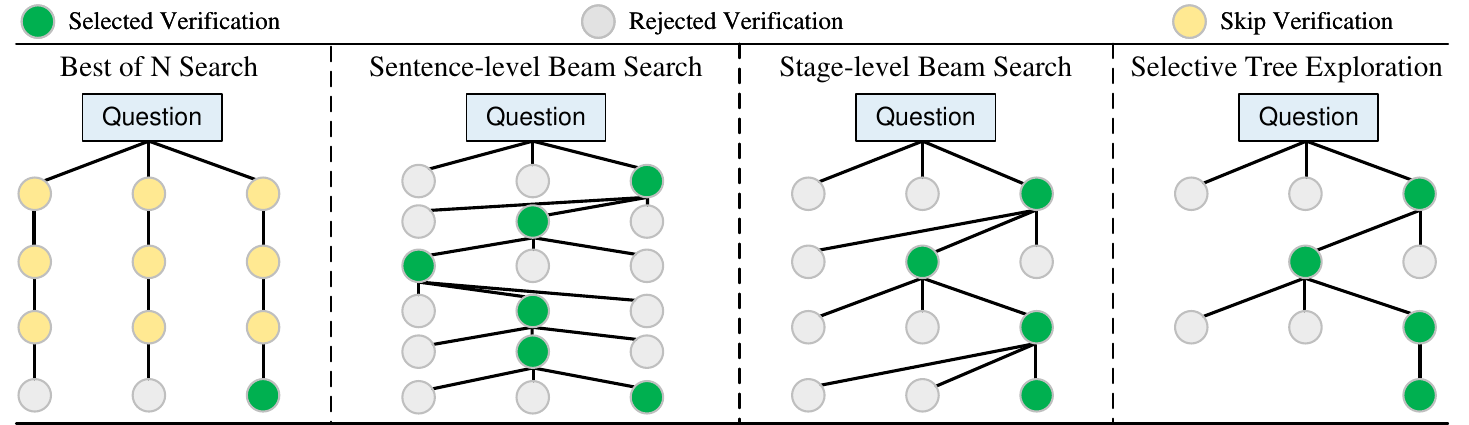} 
\caption{Solution expansion \& sampling illustration. Best-of-N search generates N complete responses and selects the best one; Sentence-level Beam Search generates multiple candidates for each sentence and selects the best one; Similarly, Stage-level Beam Search generates multiple candidates for each reasoning step and selects the best one. In contrast, our Selective Tree Exploration dynamically expands each reasoning step node, explores multiple reasoning steps as candidates only when necessary, and selects the best option at each step. Our method balances search performance and computational time overhead.}
\label{Selective_Tree_Exploration}
\end{figure}

\begin{figure*}[t]
\centering
\includegraphics[width=0.99\textwidth]{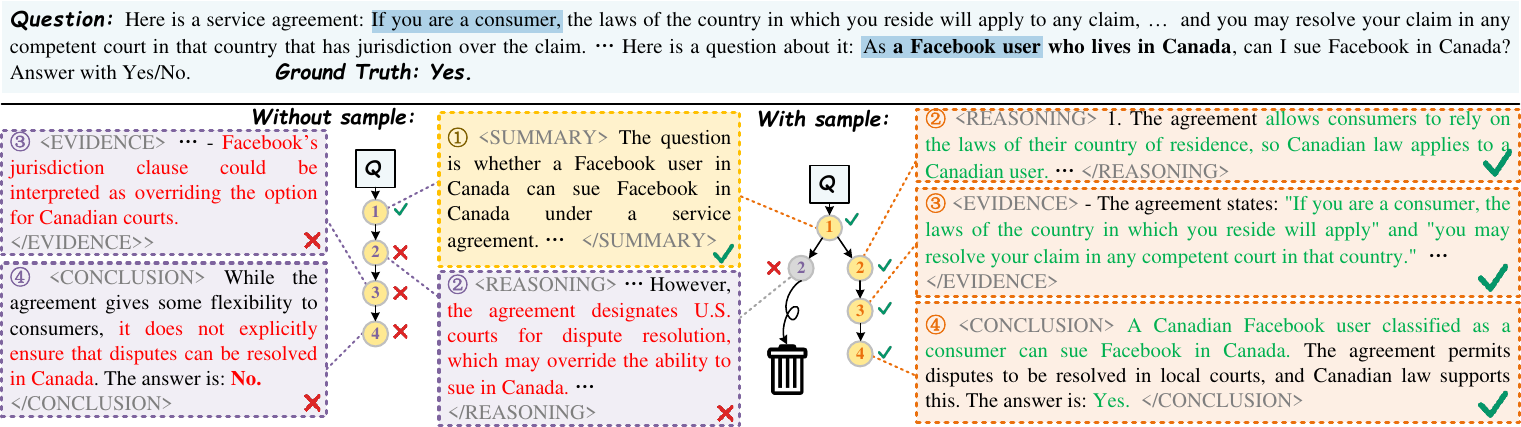} 
\caption{The role of solution expansion \& sampling. Intermediate steps in single inference (without sample) may contain errors, while solution expansion \& sampling can find better reasoning paths.}
\label{sample_exp}
\end{figure*}

When explicitly prompting GPT-4o to generate multiple structured reasoning steps, we require the model to use special tokens (e.g., <SUMMARY>) for segmentation. However, we want Domain$o1$s to organize and initiate necessary steps independently during reasoning to maintain general capabilities. Therefore, we remove all special tokens from the answers during supervised fine-tuning. After training, the model activates each reasoning step based on its own judgment.

\subsection{Solution Expansion \& Sampling}
After supervised fine-tuning, the model can output responses in CoT format. To further enhance the model's reasoning abilities, we enable the model to explore the solution space, and autonomously expand and sample reasoning paths. During sampling, we introduce $V(s_{t+1})$ to evaluate the expected return of reasoning step $s_{t+1}$. Although $V(s_{t+1})$ can be constructed through direct introduction or training of step-level reward models~\cite{chen2024alphamath,xie2024monte,xu2024llava}, this creates additional model training and inference overhead. In our implementation, we use a more direct but effective approach. We introduce the perplexity $p$ of LLM when generating $s_{t+1}$ to serve as $V(s_{t+1})$:
\vspace{-8pt}
\begin{equation}
p= \exp(-\frac{1}{N}\sum_{i=1}^{N}\log(\frac{e^{z_{i,k}}}{\sum_{j=1}^M e^{z_{i,j}}})),
\end{equation}
where $N$ represents the number of tokens in $s_{t+1}$, $z_{i,k}$ is the logit value of the actually generated token $k$ at position $i$, $z_{i,j}$ is the logit value of candidate token $j$ at position $i$, $M$ is the vocabulary size representing the number of all candidate tokens, and $\frac{e^{z_{i,k}}}{\sum_{j=1}^M e^{z_{i,j}}}$ is the softmax probability of the actually generated token. Overall, we propose Selective Tree Exploration for solution expansion \& sampling, following these phases:

\textbf{(1)} Calculate the perplexity value $p$ of tokens at the generation step.

\textbf{(2)} If $p \geq \theta$ ($\theta$ is the sampling threshold), regenerate the step until $p < \theta$ or reach the maximum regeneration count $K$ (i.e., maximum beam size). If $p$ of all $K$ generations are no less than threshold $\theta$, greedily sample the candidate with minimum $p$ from the $K$ candidates.

\textbf{(3)} Continue to generate the next step based on the selected step, repeat phases \textbf{(1)-(3)} until the complete answer is generated.

As shown in Figure~\ref{Selective_Tree_Exploration}, compared to Best-of-N Search~\cite{weng2022large,jiang2023tigerscore}, Sentence-level Beam Search~\cite{chen2024sentence} and Stage-level Beam Search~\cite{xu2024llava}, Selective Tree Exploration balances search performance and time cost. When $\theta$ is set to 0, Selective Tree Exploration becomes Stage-level Beam Search as it explores $K$ paths at each step. When $\theta$ is set to an extremely large value, Selective Tree Exploration degenerates into CoT with a single inference chain. In other cases, Selective Tree Exploration only expands reasoning paths when necessary, which reduces unnecessary overhead.

To illustrate the role of solution expansion \& sampling, as shown in Figure~\ref{sample_exp}, when inference without sampling, although the model generates the reasoning process, errors in intermediate steps (starting from <REASONING>) lead to error accumulation, ultimately resulting in incorrect results. Through exploration and expansion of solution paths, better reasoning paths can be found, leading to more accurate results.

\begin{table*}[t]
\adjustbox{max width=\textwidth}{
\begin{tabular}{>{\raggedright\arraybackslash}p{4.5cm}|c|ccccc|c|ccc|c}
\toprule
\multicolumn{1}{c|}{\multirow{2}{*}{\textbf{Model}}} & \multirow{2}{*}{\makecell{\textbf{Model}\\\textbf{size}}} & \multicolumn{5}{c|}{\textbf{Interpretation}} & \makecell{\textbf{Rule-application/}\\\textbf{Rule-conclusion}} & \multicolumn{3}{c|}{\textbf{Rhetorical-analysis}} & \multirow{2}{*}{\textbf{Avg.}} \\ \cmidrule{3-11}
\multicolumn{1}{c|}{} &  & \textbf{CC} & \textbf{CAUD} & \textbf{MAUD} & \textbf{PP} & \textbf{IP} & \textbf{PJ} & \textbf{Scalr} & \textbf{TTP} & \textbf{TTD} &  \\ 
\midrule
Qwen-2.5-Instruct & 7B & 86.36 & 80.08 & 78.75 & 52.38 & 48.12 & 64.00 & 78.98 & 99.07 & 76.96 & 73.86 \\
Llama-3-Instruct & 8B & 85.86 & 81.20 & 67.43 & 61.63 & 50.37 & 54.00 & 75.83 & \textbf{100.00} & 78.18 & 72.72 \\
OpenO1-Llama & 8B & 85.10 & 81.31 & 74.54 & 62.36 & 50.37 & 60.00 & 80.03 & 91.52 & 77.58 & 73.65 \\
OpenO1-Qwen & 7B & 84.85 & 80.13 & 79.11 & 59.27 & 48.87 & 66.00 & 80.38 & 88.78 & 76.64 & 73.78 \\
Open-Australian-Legal & 1.5B & 0.00 & 0.00 & 1.20 & 17.64 & 1.50 & 22.00 & 0.00 & 0.00 & 0.00 & 4.70 \\
DISC-LawLLM & 13B & 50.00 & 32.98 & 64.77 & 48.09 & 19.55 & 56.00 & 70.05 & 5.60 & 20.60 & 40.85 \\
Law-LLM & 7B & 10.86 & 1.59 & 30.87 & 3.05 & 2.26 & 0.00 & 58.49 & 8.41 & 13.33 & 14.32 \\
Law-Chat & 7B & 80.30 & \textbf{82.31} & 39.75 & 51.69 & 33.83 & 48.00 & 76.36 & 54.21 & 52.73 & 57.69 \\
Lawma & 8B & 47.73 & 34.14 & 69.93 & 53.31 & 47.37 & 36.00 & 78.46 & 6.54 & 26.67 & 44.46 \\
\cmidrule{1-12}
\textbf{Domain-CoT-legal} & 7B & 87.88 & 80.59 & \textbf{80.47} & 65.81 & 50.37 & 70.00 & 86.69 & 94.40 & 77.58 & 77.09 \\
\textbf{Domain$o1$s-legal} & 7B & \textbf{88.64} & 81.76 & 80.33 & \textbf{66.54} & \textbf{52.63} & \textbf{72.00} & \textbf{88.97} & 95.33 & \textbf{78.78} & \textbf{78.33} \\
\bottomrule
\end{tabular}
}
\caption{Model accuracy (\%) on legal reasoning QA tasks. Avg. represents the mean accuracy across all tasks.}
\label{tab_legal_main}
\end{table*}

\section{Experiments}
In this section, we evaluate the performance of Domain$o1$s on stock investment recommendation and legal reasoning QA tasks. Our work aims to address the following questions:
\textbf{RQ1}: How does Domain$o1$s perform in answer accuracy compared to other LLM methods?
\textbf{RQ2}: What are the limitations of accuracy-based evaluation metrics in domain tasks, and how can we better evaluate model performance?
\textbf{RQ3}: How do fine-tuning and solution expansion \& sampling help improve the performance of Domain$o1$s?
\subsection{Experimental Settings}
\textbf{Baselines.} To validate Domain$o1$s's performance on high-stakes domain tasks, we compare it with general purpose LLMs and domain LLMs trained or fine-tuned with domain data. 

General Purpose LLMs: We choose Qwen-2.5-Instruct~\cite{qwen2.5} and Llama-3-Instruct~\cite{llama3modelcard} as general purpose LLM baselines due to their remarkable performance on many downstream tasks. We also select OpenO1-Llama and OpenO1-Qwen~\cite{openO1team2024} as representatives of o1-type model baselines.

Financial Domain LLMs: Finance-LLM~\cite{cheng2024adapting},
Finance-Chat~\cite{cheng2024adapting},
Finance-Llama-3~\cite{cheng2024instruction},
FinGPT-Forecaster~\cite{yang2023fingpt},
Llama-2-taiwan-btc~\cite{lanz2024llama2}, and
SEP~\cite{koa2024learning}.

Legal Domain LLMs:
Open-Australian-Legal-LLM~\cite{butler-2023-open-australian-legal-llm},
DISC-LawLLM~\cite{yue2023disc},
Law-LLM~\cite{cheng2024adapting},
Law-Chat~\cite{cheng2024adapting}, and
Lawma~\cite{dominguez2024lawma}.

\textbf{Datasets.} For the stock investment recommendation task, we select the stock prediction dataset provided by Koa et al.~\cite{koa2024learning}. This dataset contains price data and tweet information for the top 5 stocks from 11 industries during 2020-2022, comprising 7,866 test question entries. The task is constructed to predict whether a stock will rise or fall on the next trading day based on facts contained in tweets from the previous 5 days. Any neutral answers are considered incorrect. Due to the large volume of daily tweets, we fine-tune Qwen-2.5-Instruct~\cite{qwen2.5} to generate daily tweet summaries and apply these summaries as input for all models.

For the legal reasoning QA task, we select LegalBench~\cite{guha2024legalbench}, a dataset composed of numerous legal QA datasets and benchmarks. LegalBench includes 5 categories of legal tasks. We select three reasoning-related categories: Rule-application/Rule-conclusion, Interpretation, and Rhetorical-understanding, encompassing 9 datasets with a total of 35,053 test questions. Question types include true/false and multiple-choice questions.

\textbf{Implementation Details.} In this work, our Domain$o1$s is developed based on Qwen-2.5-Instruct~\cite{qwen2.5}. During the fine-tuning phase for enhancing reasoning capabilities, we set the learning rate, epoch, batch size, gradient accumulation, and maximum tokens length to 5e-5, 120, 2, 2, and 2048 respectively. The $\theta$ and $K$ in the sampling process are set to 1.1 and 2 respectively. The experimental hardware, software, and other configuration details can be found in Appendix~\ref{Appendix_Experimental_Setup}.

\begin{table}[t]
\resizebox{\columnwidth}{!}{
\begin{tabular}{>{\raggedright\arraybackslash}p{3.5cm}ccc}
\toprule
\multicolumn{1}{c}{\textbf{Model}} & \textbf{Model Size} & \textbf{Accuracy} & \textbf{MCC} \\
\midrule
Qwen-2.5-Instruct & 7B & 51.18 & -0.017 \\
Llama-3-Instruct & 8B & 51.41 & 0.017 \\
OpenO1-Llama & 8B & 50.87 & 0.014 \\
OpenO1-Qwen & 7B & 51.02 & 0.010 \\
Finance-LLM & 7B & 48.05 & -0.075 \\
Finance-Chat & 8B & 47.16 & -0.004 \\
Finance-Llama-3 & 8B & 49.03 & -0.047 \\
FinGPT & 7B & 46.13 & 0.016 \\
Llama-2-taiwan-btc & 7B & 50.66 & -0.002 \\
SEP & 7B & 48.35 & 0.018 \\
\cmidrule{1-4}
\textbf{Domain-CoT-finance} & 7B & 51.52 & 0.020 \\
\textbf{Domain$o1$s-finance} & 7B & \textbf{51.98} & \textbf{0.021} \\
\bottomrule
\end{tabular}
}
\caption{Model accuracy (\%) and MCC on stock investment recommendation tasks.}
\label{tab_stock_main}
\end{table}

\subsection{Prediction Performance (RQ1)}
In this section, we compare Domain$o1$s with relevant baselines to evaluate the answer accuracy.

Table~\ref{tab_legal_main} and Table~\ref{tab_stock_main} report the quantitative results for legal reasoning QA and stock investment recommendations tasks respectively. For all models where answers cannot be directly parsed from responses, we use GPT-3.5-turbo-16k~\cite{ouyang2022training} to extract the chosen options from responses for fair comparison. Additionally, given that not all stock price movements are necessarily caused by the provided text, accuracy results may not fully indicate a model's reasoning capabilities, as they include some random guesses for non-informative text~\cite{koa2024learning}. Following stock prediction research~\cite{ding2015deep,feng2018enhancing}, we also calculate the Matthews Correlation Coefficient (MCC) as an evaluation metric, which considers the ratios of true and false positives and negatives~\cite{chicco2020advantages,chicco2021matthews}. We observe that Domain$o1$s outperforms its base model Qwen-2.5-Instruct on almost all tasks, despite being fine-tuned on only a small amount of data. Moreover, Domain$o1$s and Domain-CoT (model with reasoning-enhanced fine-tuning, without solution expansion \& sampling) achieve the best accuracy or MCC on nearly all tasks, even surpassing LLMs that are carefully designed and trained on domain datasets, especially on legal reasoning tasks as shown in Table~\ref{tab_legal_main}. Although these legal LLMs learn domain knowledge through pre-training or fine-tuning, they lack the reasoning capability to derive correct answers, in contrast to our models. We also analyze the reasoning chain length and inference time of Domain$o1$s and baselines, see Appendix~\ref{time_length}.

\subsection{Explainability Evaluation Pipeline (RQ2)}
\label{section:rq2}
In previous research, most domain tasks use accuracy as the primary evaluation metric~\cite{koa2024learning,yang2022numhtml,guha2024legalbench}. This evaluation metric makes it difficult to distinguish between models that truly understand and reasonably utilize context and those that simply rely on partial text or overfit on pre-trained domain knowledge~\cite{zhang2024uncovering,bordt2404elephants}. We sample two subsets from the test sets of stock investment recommendation and legal reasoning QA, with details available in Appendix~\ref{Appendix_Sub-datasets}.

\begin{figure}[t]
\centering
\includegraphics[width=0.49\textwidth]{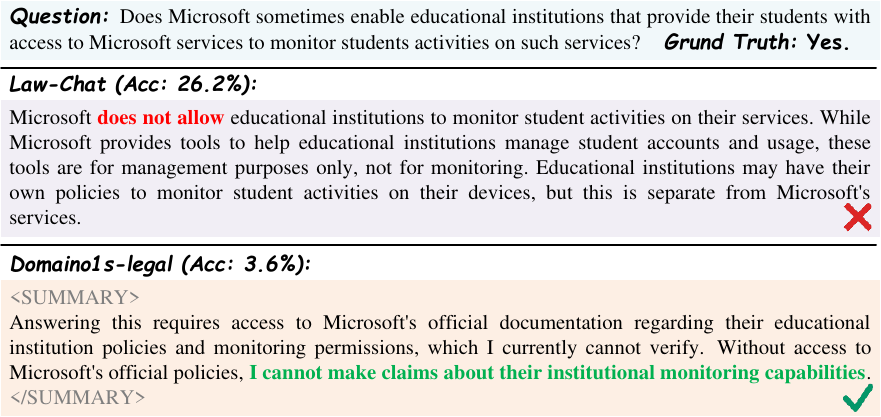} 
\caption{Models responses and accuracy on legal reasoning QA tasks after removing key conditions.}
\label{sub_legal}
\end{figure}

In the stock investment recommendation task, stock tweets are manually classified into Positive and Negative tweets and combined in different ratios as model inputs. We compare Domain$o1$s-finance with Finance-Llama-3. As shown in Table~\ref{tab_sub_stock}, when the Positive:Negative ratio of tweets is 0.5:0.5, models' responses maintain a similar 1:1 ratio between Positive and Negative predictions. However, when either Positive or Negative tweets dominate the input, Finance-Llama-3 typically ignores tweets with the opposite sentiment and bases its answer solely on the majority sentiment. In contrast, Domain$o1$s-finance still considers the minority sentiment tweets and generates answers by comprehensively evaluating all tweets. However, both models achieve similar accuracy, making it challenging to determine through accuracy metrics alone whether the models truly understand and reasonably utilize the context in the inputs, rather than overfitting or hallucinating. For the legal reasoning task subset, key conditions are removed from the question text, making it impossible to answer the tasks correctly. As shown in Figure~\ref{sub_legal}, although Law-Chat achieves higher accuracy than Domain$o1$s-legal, its answers are mostly random responses generated from overfitted legal knowledge, while Domain$o1$s-legal refuses to answer due to the absence of necessary reasoning conditions, resulting in an accuracy close to 0. This indicates that accuracy alone is insufficient to determine whether models blindly overfit using domain knowledge to generate irrelevant answers.

\begin{table}[t]
\centering
\small
\setlength{\tabcolsep}{3pt}  
\begin{tabular*}{\columnwidth}{@{\extracolsep{\fill}}cc|ccc|ccc@{}}
\toprule
\multicolumn{2}{c}{\textbf{Model}}    & \multicolumn{3}{|c}{\textbf{Finance-Llama-3}}         & \multicolumn{3}{|c}{\textbf{Domain$o1$s-finance}}    \\
\midrule
\multicolumn{2}{c}{\textbf{Tweets}}   & \multicolumn{6}{|c}{\textbf{Response}}                                                                       \\
\midrule
\textbf{Pos.} & \textbf{Neg.} & \textbf{Pos.} & \textbf{Neg.} & \textbf{Acc} & \textbf{Pos.} & \textbf{Neg.} & \textbf{Acc} \\  
\midrule
0.1          & 0.9          & 0.018        & 0.982        & 50.45        & 0.173        & 0.827        & 50.91        \\
0.3          & 0.7          & 0.182        & 0.818        & 48.64        & 0.391        & 0.609        & 49.55        \\
0.5          & 0.5          & 0.545        & 0.455        & 48.18        & 0.527        & 0.473        & 49.09        \\
0.7          & 0.3          & 0.882        & 0.118        & 51.36        & 0.777        & 0.223        & 50.91        \\
0.9          & 0.1          & 1.000         & 0.000         & 50.00        & 0.882        & 0.118        & 50.45        \\
\bottomrule
\end{tabular*}
\caption{Model's accuracy and prediction ratios for Positive (Pos.) and Negative (Neg.) of the stock trend under varying proportions of Pos. and Neg. tweets in stock investment prediction tasks.}
\label{tab_sub_stock}
\end{table}

\begin{table}[t]
\resizebox{\columnwidth}{!}{
\begin{tabular}{lc|cc}
\toprule
\multicolumn{2}{c}{\textbf{Stock}} & \multicolumn{2}{|c}{\textbf{Legal}} \\
\midrule
Qwen-2.5-Instruct & 6.281 & Qwen-2.5-Instruct & 3.428 \\
Llama-3-Instruct & 6.129 & Llama-3-Instruct & 3.417 \\
OpenO1-Llama & 6.212 & OpenO1-Llama & 6.554 \\
OpenO1-Qwen & 6.227 & OpenO1-Qwen & 6.588 \\
Finance-LLM & 6.023 & Open-Australian-Legal & 5.152 \\
Finance-Chat & 5.583 & DISC-LawLLM & 0 \\
Finance-Llama-3 & 5.965 & Law-LLM & 3.838 \\
FinGPT & 3.413 & Law-Chat & 3.339 \\
Llama-2-taiwan-btc & 0 & Lawma & 0 \\
SEP & 6.182 & & \\
\cmidrule{1-4}
\textbf{Domain$o1$s-finance} & \textbf{6.359} & \textbf{Domain$o1$s-legal} & \textbf{6.677} \\
\bottomrule
\end{tabular}
}
\caption{Comparison of explanation quality (PROOF-Score) between Domain$o1$s and baselines. For models that generate responses containing no explanations, their PROOF-Scores are set to 0.}
\label{tab_reasoning}
\end{table}

For high-stakes domain tasks such as stock investment recommendations and legal reasoning QA, non-transparent text comprehension or inadequate reasoning processes may lead to wrong conclusions or generate advice that violates ethical or legal principles. To evaluate the explainability of domain model responses, we propose a new evaluation metric called PROOF-Score (\underline{P}rincipled \underline{r}ating for reas\underline{o}ning completeness, d\underline{o}main safety, and \underline{f}actual accuracy). PROOF-Score uses GPT-4o~\cite{hurst2024gpt} to generate a score from 1 to 7 for response, considering three aspects:

• \textbf{Reasoning Completeness (RC)}: Evaluates the completeness and logical coherence. 

• \textbf{Domain Safety (DS)}: Measures the safety and appropriateness in specific domains.

• \textbf{Factual Accuracy (FA)}: Evaluates the factual accuracy of statements.

Detailed prompts can be seen in Appendix~\ref{Appendix_PROOF-Score_Generation}. Here, we define:
\begin{equation}
\label{eq:proofscore}
\text{PROOF-Score} = \frac{RC + DS + FA} {3}.
\end{equation}

Table~\ref{tab_reasoning} shows PROOF-Scores of models on two tasks. Domain$o1$s achieves the highest scores on both tasks, even though we do not train specifically for these three metrics. This indicates Domain$o1$s can inherently consider these factors to generate better responses. We also observe that even when a model's response is incorrect in terms of results, GPT-4o may still give a high PROOF-Score because these responses contain clear and reasonable logic. This may be inappropriate for tasks requiring strict accuracy, where prediction accuracy should be considered the primary metric. However, for tasks lacking standard answers or without unique correct answers (e.g., long-term investment advice, asset allocation recommendations), using PROOF-Score becomes effective in evaluating the explainability of model responses.

\begin{table}[t]
\centering
\small
\begin{tabular*}{\columnwidth}{@{\extracolsep{\fill}}lcc@{}}
\toprule
\textbf{Method} & \textbf{Acc} & \textbf{time(s)} \\
\midrule
w/o Sample & 86.69 & 8.35 \\
\midrule
Best-of-N Search & 87.56 & 40.26 \\
Sentence-level Beam Search & 84.93 & 334.20 \\
Stage-level Beam Search & 88.44 & 133.68 \\
\cmidrule{1-3}
\textbf{Selective Tree Exploration} & \textbf{89.14} & \textbf{15.18} \\
\bottomrule
\end{tabular*}
\caption{Accuracy (\%) and average inference time comparison between our Selective Tree Exploration and other search methods on the Scalr dataset. Our method (with $\theta=1.1$) outperforms other approaches under the same beam size settings.}
\label{tab_sample}
\end{table}

\begin{table}[t]
\centering
\small
\begin{tabular*}{\columnwidth}{@{\extracolsep{\fill}}lccc@{}}
\toprule
\textbf{Method} & \textbf{$K$} & \textbf{Acc} & \textbf{time(s)} \\
\midrule
w/o Sample & 1 & 86.69 & 8.35 \\
\cmidrule{1-4}
\multirow{3}{*}{\textbf{Selective Tree}} & 2 & 88.97 & \textbf{24.88} \\
\multirow{3}{*}{\textbf{Exploration}} & 3 & 89.14 & 45.77 \\
& 4 & 89.84 & 72.55 \\
& 5 & \textbf{90.01} & 93.95 \\
\bottomrule
\end{tabular*}
\caption{Accuracy (\%) and average inference time of Domain$o1$s-legal on the Scalr dataset under different beam size $K$ settings. $\theta$ is set to 1.05.}
\label{tab_K}
\end{table}

\begin{table}[t]
\centering
\small
\begin{tabular*}{\columnwidth}{@{\extracolsep{\fill}}lccc@{}}
\toprule
\textbf{Method} & \textbf{$\theta$} & \textbf{Acc} & \textbf{time(s)} \\
\midrule
w/o Sample & 10000 & 86.69 & 8.35 \\
\cmidrule{1-4}
\multirow{5}{*}{\textbf{Selective Tree Exploration}} & 1.4 & 87.21 & \textbf{8.40} \\
& 1.3 & 87.91 & 8.44 \\
& 1.2 & 88.97 & 10.16 \\
& 1.1 & 89.14 & 15.18 \\
& 1.0 & \textbf{89.49} & 51.03 \\
\bottomrule
\end{tabular*}
\caption{Accuracy (\%) and average inference time of Domain$o1$s-legal on the Scalr dataset under different sampling threshold $\theta$ settings. $K$ is set to 3.}
\label{tab_theta}
\end{table}

\subsection{Ablation Study (RQ3)}
In this section, we evaluate the impact of fine-tuning and solution expansion \& sampling on Domain$o1$s's performance. We primarily focus on accuracy metrics in this section, while presenting explainability analysis in Appendix~\ref{Appendix_Explainability}.

\textbf{Enhancing Reasoning Fine-tuning}. As shown in Table~\ref{tab_legal_main} and Table~\ref{tab_stock_main}, Domain-CoT represents the model configuration using only reasoning-enhanced fine-tuning without solution expansion \& sampling. Compared to the base model Qwen-2.5-Instruct, Domain-CoT achieves performance improvements on almost all datasets, which demonstrates that reasoning-enhanced fine-tuning improves the model's reasoning capabilities on domain tasks.

\textbf{Solution Expansion \& Sampling}. Table~\ref{tab_sample} shows the performance comparison on Scalr (a dataset in LegalBench) between best-of-N search~\cite{weng2022large,jiang2023tigerscore}, Sentence-level Beam Search ~\cite{chen2024sentence}, Stage-level Beam Search~\cite{xu2024llava}, and our Selective Tree Exploration. The baseline search methods use the setup from Xu et al.~\cite{xu2024llava}, which uses the policy model to evaluate the relative quality of reasoning chains or steps, in contrast to our perplexity-based approach. Results demonstrate that under the same beam setting of $K=3$, Selective Tree Exploration achieves comparable or better performance compared to all baseline approaches (with and without search) while requiring less computational time for inference than other search methods.

To better illustrate the effectiveness of our Selective Tree Exploration as exploration paths increase, we evaluate model performance under different settings of $K$ and $\theta$ on the Scalr dataset. As shown in Table~\ref{tab_K}, using Selective Tree Exploration brings performance improvements compared to methods without sampling ($K=1$). Model accuracy improves as $K$ increases, indicating that our Selective Tree Exploration is scalable. As shown in Table~\ref{tab_theta}, model accuracy improves as $\theta$ decreases, as this similarly expands the paths explored by Selective Tree Exploration. However, both increasing $K$ and decreasing $\theta$ lead to longer inference time. Due to computational resource constraints, we only set $K=2$, $\theta=1.1$. However, we demonstrate that increasing beam size $K$ and decreasing sampling threshold $\theta$ will lead to performance improvements.

\section{Acknowledgment} This work is supported by the National Key R\&D Program of China [2022YFF0902703].

\section{Conclusion \& Future Works}
In this work, we introduce Domain$o1$s and its two model variants for finance and legal domains, guiding LLMs towards explainable high-stakes domain answers. We construct two datasets to fine-tune Qwen-2.5-Instruct and propose Selective Tree Exploration for enabling LLMs to perform multi-stage reasoning. The superior performance on datasets demonstrates Domain$o1$s's exceptional potential in high-stakes domains.

In future work, we plan to build larger training datasets to enhance domain models' reasoning abilities. We also plan to create Domain$o1$s variants using domain-specific pre-trained base models to better solve tasks requiring domain expertise.

\section{Limitations}
Despite the promising results achieved by Domain$o1$s, there are some limitations. First, while our Selective Tree Exploration method effectively balances search performance and computational costs, the additional inference time required for tree exploration may impact the model's real-time application scenarios, such as in situations requiring high response speed. Second, although we construct high-quality CoT datasets using GPT-4o, the relatively small size of training data (2,000 examples each for finance and legal domains) may limit the model's ability to handle extremely rare or complex domain-specific cases. Additionally, while PROOF-Score provides a comprehensive evaluation framework, research on using LLMs as judges suggests that further refinement and elaboration of evaluation metrics may be beneficial~\cite{gu2024survey}. Finally, our current implementation focuses on stock recommendation and legal reasoning tasks, and the generalizability of our approach to other domain applications requires further investigation. These limitations point to promising directions for future research, such as optimizing inference efficiency, expanding training datasets, and extending the framework to broader domain applications.

\section{Ethical Considerations}

\subsection{Fairness and Accessibility}

We recognize that the computational resources required for training and inference of large language models (LLMs) and tree search exploration may limit accessibility for researchers and practitioners with fewer resources. To address this, we will open-source our implementation and provide efficient variants that can run on consumer-grade hardware. Additionally, we will release the training datasets (CoT-stock-2k and CoT-legal-2k) to enhance reproducibility and facilitate broader participation in this research direction.

\subsection{Potential Risks in Financial and Legal Applications}

For financial applications, we acknowledge that Domain$o1$s-finance's advice, while explainable, should be viewed as restricted investment references. To mitigate potential risks:

\begin{itemize}[nosep]
\item We explicitly state that Domain$o1$s-finance's outputs should serve as one of many considerations when users make actual investment decisions.
\item We implement safety checks in the Domain Safety (DS) metric of PROOF-Score to detect potentially harmful or high-stake advice.
\item We emphasize the importance of human oversight and professional judgment in interpreting model reasoning.
\end{itemize}

For legal applications, Domain$o1$s-legal is intended to assist rather than replace legal professionals. To mitigate potential risks:

\begin{itemize}[nosep]
\item We explicitly state that Domain$o1$s-legal is proposed as a support tool rather than a substitute for professional legal advice.
\item We detect responses that contradict legal facts by evaluating the Factual Accuracy (FA) metric of PROOF-Score.
\item We emphasize the importance of human oversight and professional judgment in interpreting model reasoning.
\end{itemize}

\subsection{Privacy and Data Security}

We have taken multiple measures to protect privacy and ensure data security:

\begin{itemize}[nosep]
\item Our datasets have been carefully screened and curated to exclude sensitive personal information.
\item The model's inference process is designed to focus on public information.
\item Implement rate limiting and access controls after model and dataset open-sourcing to prevent potential misuse.
\end{itemize}

\subsection{Environmental Impact}

We acknowledge the environmental impact of training and running large language models. To minimize this:

\begin{itemize}[nosep]
\item Our proposed Selective Tree Exploration method is designed to improve computational efficiency and reduce inference overhead.
\item We provide guidance on optimal hyperparameter settings and encourage the selection of hyperparameter configurations that balance computational costs with model performance to reduce unnecessary computation.
\end{itemize}

Through these considerations and safeguards, we aim to ensure Domain$o1$s makes positive contributions to the field while minimizing potential risks and negative impacts. We encourage ongoing dialogue with stakeholders and welcome community feedback to further improve the ethical implementation of our technology.

\bibliography{custom}

\appendix
\section{Experimental Setup}
\label{Appendix_Experimental_Setup}
All experiments are conducted using an AMD EPYC 7H12 64-Core processor as CPU and four 48GB NVIDIA RTX 6000 Ada GPUs. For each variant of Domain$o1$s, fine-tuning takes approximately 48 GPU hours per run. The system environment uses CUDA version 12.4, Python version 3.10.15, PyTorch version 2.5.1, and transformers 
version 4.45.2. The random seed is set to 42.

We employ LoRA (Low-Rank Adaptation) for fine-tuning. The base model is Qwen2.5-7B-Instruct. We use the qwen template with Flash Attention enabled. The training dataset is preprocessed using 16 workers with a maximum sequence length of 2,048 tokens.

The LoRA hyperparameters are set as follows: rank = 8, alpha = 16, and dropout = 0, targeting all model layers. For optimization, we use the AdamW optimizer with a learning rate of 5e-5 and cosine learning rate scheduling. The training runs for 120 epochs. We employ mixed-precision training using bfloat16 format.

The batch size is set to 2 per device with a gradient accumulation of 2 steps, effectively creating a batch size of 16 (2 × 2 × 4 GPUs). Gradient clipping is applied with a maximum norm of 1.0. The model checkpoints are saved every 100 steps, with loss logging occurring every 5 steps. 

\section{CoT Data Generation}
\label{Appendix_CoT_Data_Generation}
\begin{figure}[htbp]
\centering
\includegraphics[width=0.49\textwidth]{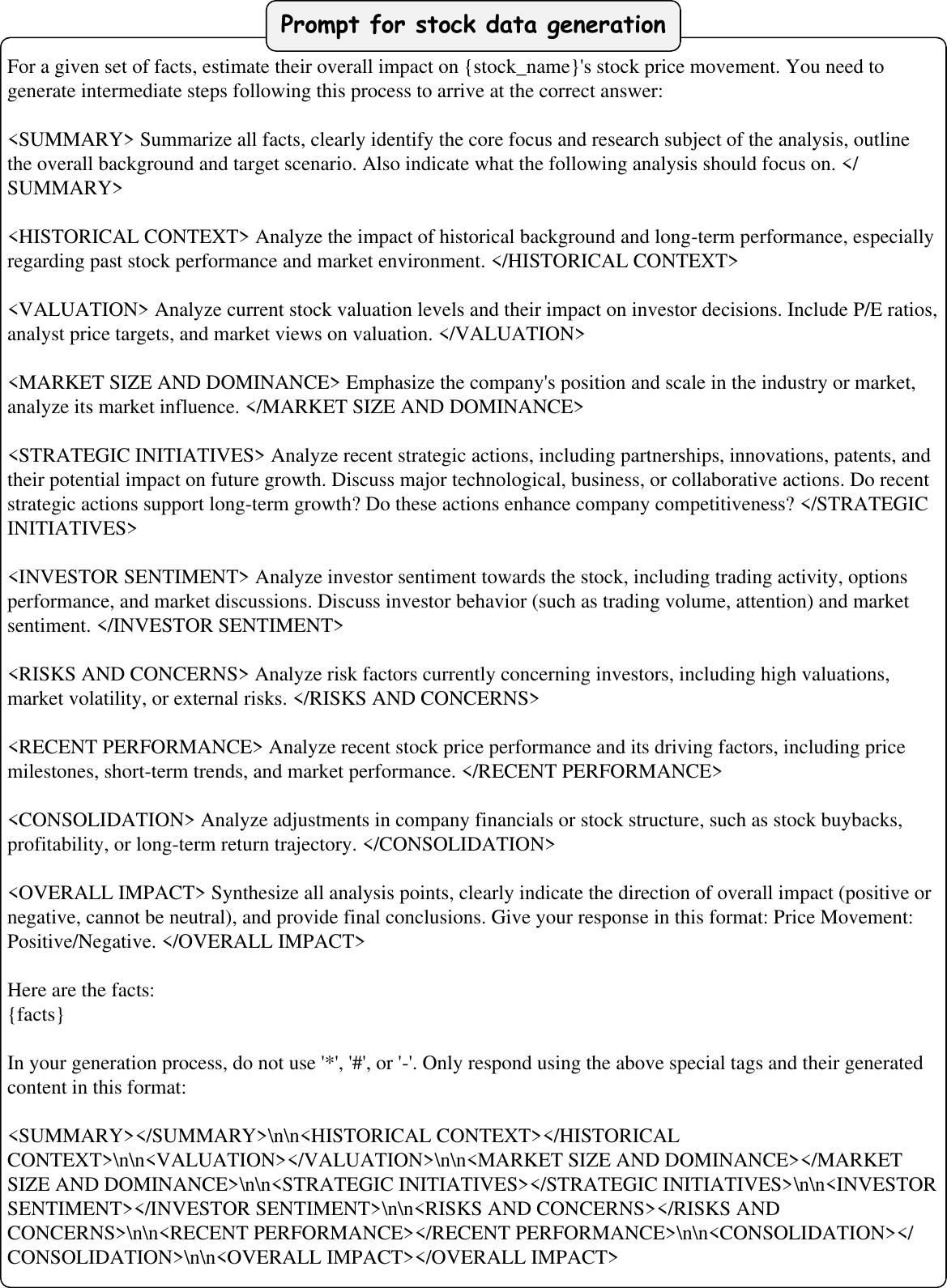}
\caption{Prompt template for stock investment recommendation.}
\label{fig:prompt_stock}
\end{figure}

\begin{figure}[htbp]
\centering
\includegraphics[width=0.49\textwidth]{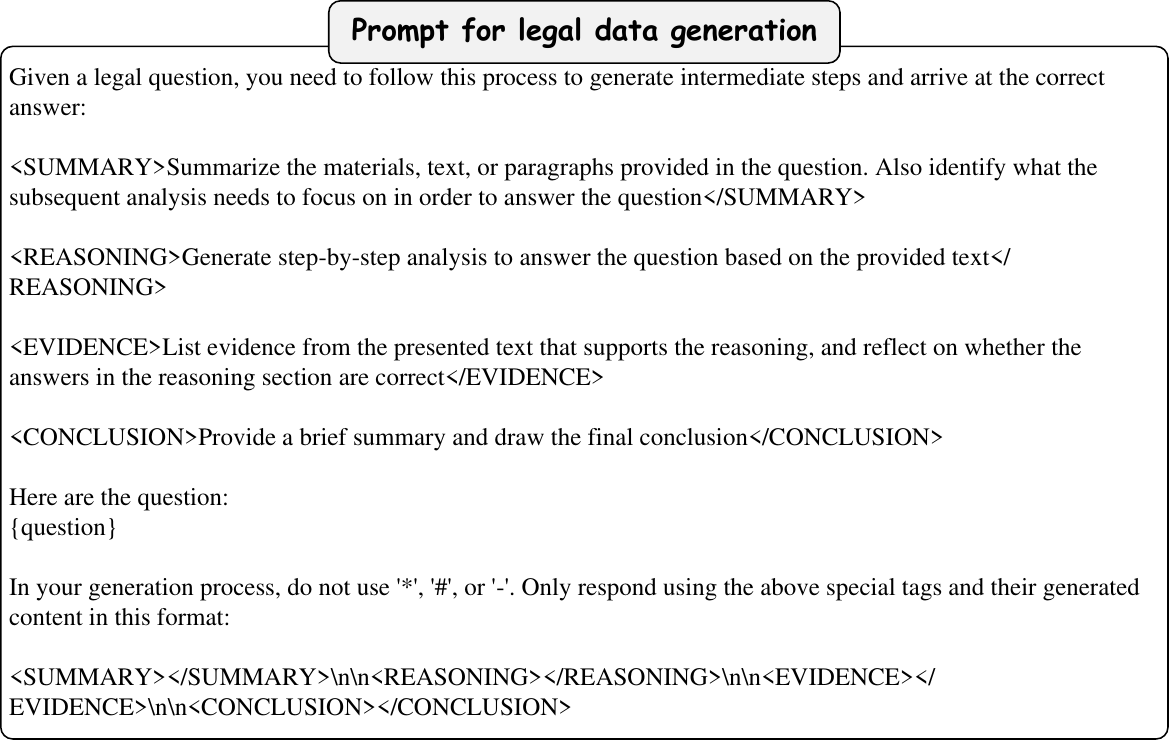}
\caption{Prompt template for legal reasoning QA.}
\label{fig:prompt_legal}
\end{figure}

Figure~\ref{fig:prompt_stock} and Figure~\ref{fig:prompt_legal} are the prompt templates for instructing GPT-4o to generate responses in CoT format.

\begin{figure}[htbp]
\centering
\includegraphics[width=0.49\textwidth]{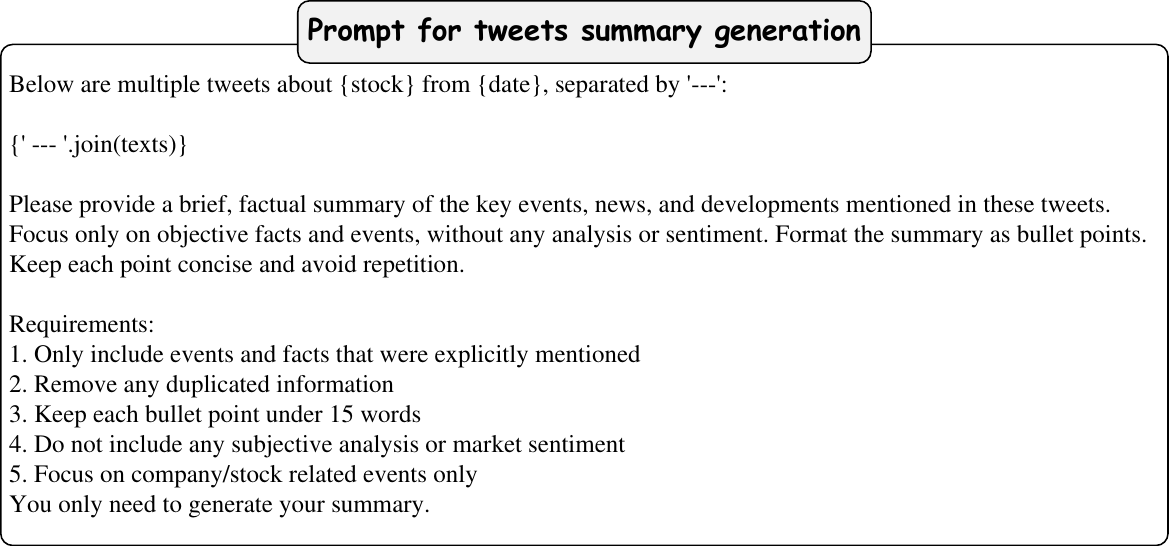}
\caption{Prompt template for tweet summarization.}
\label{fig:prompt_summary}
\end{figure}

Figure~\ref{fig:prompt_summary} is the prompt template for instructing GPT-4o and Qwen-2.5-Instruct to generate tweet summaries.

\section{Answer Demonstration}
\label{Appendix_Answer_Demonstration}
\begin{figure*}[htbp]
\centering
\includegraphics[width=0.99\textwidth]{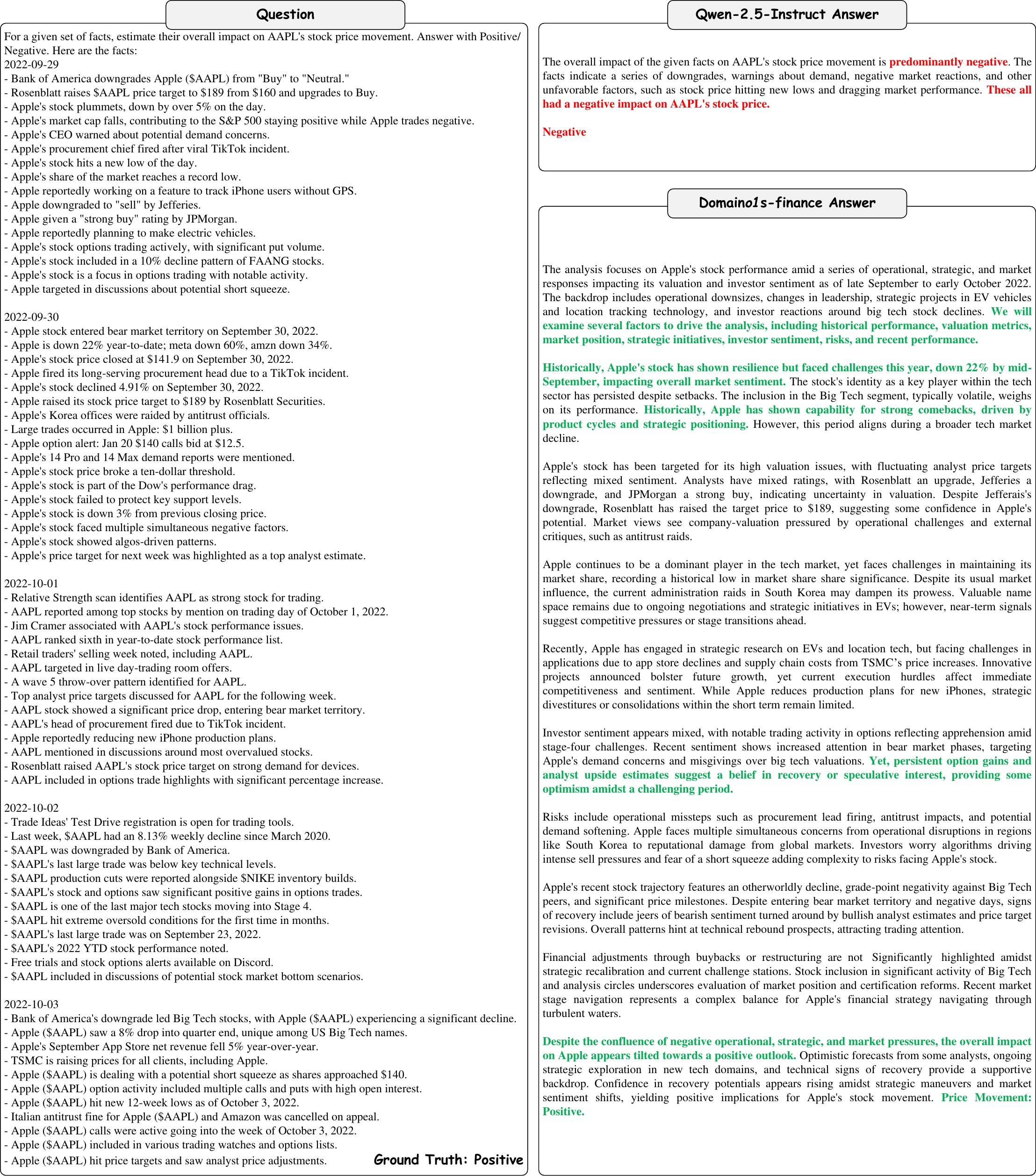}
\caption{Questions and answer demonstrations for Domain$o1$s-finance and base model.}
\label{full-exp-stock}
\end{figure*}

\begin{figure*}[htbp]
\centering
\includegraphics[width=0.99\textwidth]{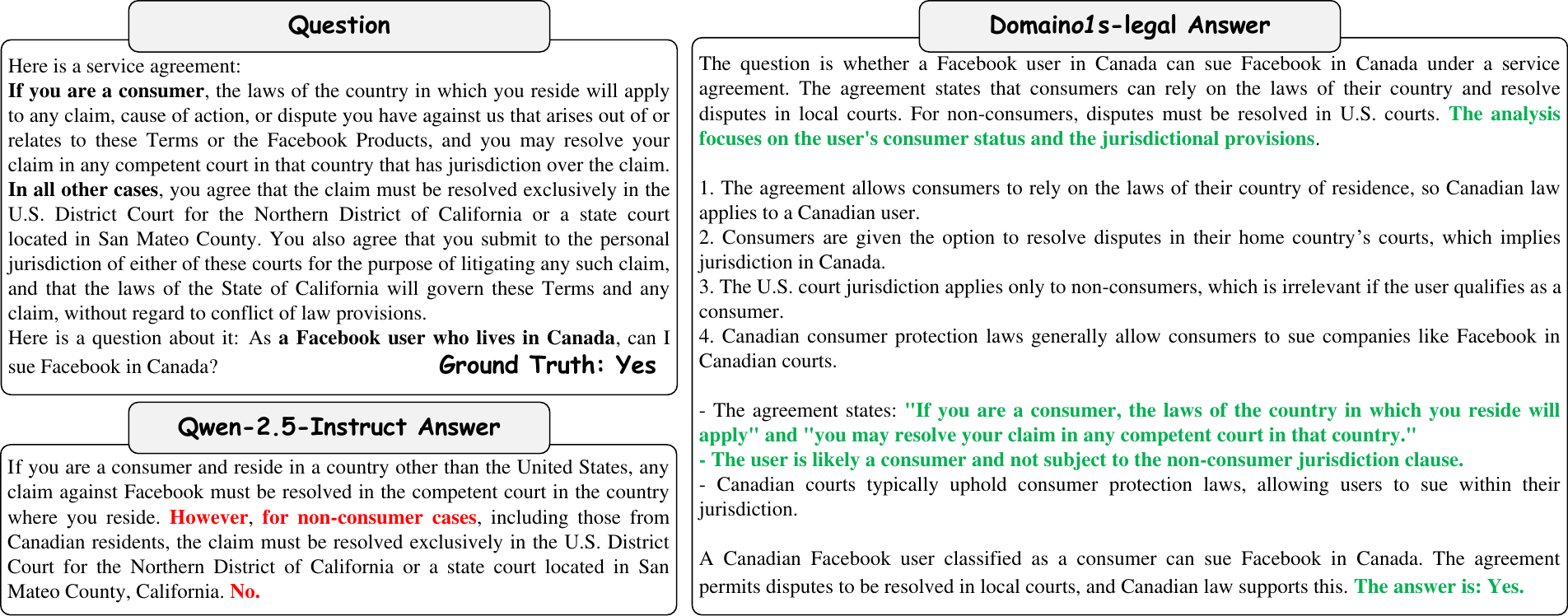}
\caption{Questions and answer demonstrations for Domain$o1$s-legal and base model.}
\label{full-exp-legal}
\end{figure*}

Figures \ref{full-exp-stock} and \ref{full-exp-legal} demonstrate complete question-answering examples for stock investment recommendation and legal reasoning QA tasks using Domain$o1$s and the base model Qwen-2.5-Instruct. Domain$o1$s does not explicitly output special tokens (e.g., <SUMMARY>), but reason according to the structured reasoning process constructed in the CoT-stock-2k and CoT-legal-2k datasets.

As shown in Figure~\ref{full-exp-stock}, Qwen-2.5-Instruct reaches an incorrect answer by focusing only on partial information (the Negative parts) while ignoring the overall context. In contrast, Domain$o1$s-finance comprehensively considers both Positive and Negative facts to draw conclusions. As shown in Figure~\ref{full-exp-legal}, Qwen-2.5-Instruct starts making reasoning errors after generating "However," incorrectly classifying the user as a non-consumer, leading to an incorrect result. In comparison, Domain$o1$s-legal avoids errors through structured reasoning paths and tree search.

\begin{table}[t]
\small
\resizebox{\columnwidth}{!}{
\begin{tabular}{>{\raggedright\arraybackslash}p{3.5cm}|c|c}
\toprule
\multicolumn{1}{c|}{\textbf{Model}} & \textbf{time(s)} & \textbf{Length} \\
\midrule
Qwen-2.5-Instruct & 6.91 & 166.2 \\
Llama-3-Instruct & \textbf{5.08} & 95.2 \\
OpenO1-Llama & 20.27 & 454.8 \\
OpenO1-Qwen & 21.09 & 465.1 \\
Finance-LLM & 14.70 & 131.3 \\
Finance-Chat & 14.52 & 130.8 \\
Finance-Llama-3 & 5.34 & 13.5 \\
FinGPT & 6.29 & 14.2 \\
Llama-2-taiwan-btc & 13.63 & 41.3 \\
SEP & 13.182 & 119.6 \\
\cmidrule{1-3}
\textbf{Domain-CoT-finance} & 18.37 & \textbf{512.1} \\
\textbf{Domain$o1$s-finance} & 27.38 & 509.8 \\
\bottomrule
\end{tabular}
}
\caption{Inference time and reasoning chain length on stock investment recommendation tasks.}
\label{tab:time_length_stock}
\end{table}

\begin{table}[t]
\small
\adjustbox{max width=\columnwidth}{
\begin{tabular}{>{\raggedright\arraybackslash}p{4.5cm}|c|c}
\toprule
\multicolumn{1}{c|}{\textbf{Model}} & \textbf{time(s)} & \textbf{Length} \\
\midrule
Qwen-2.5-Instruct & 0.65 & 1.3 \\
Llama-3-Instruct & 0.71 & 1.5 \\
OpenO1-Llama & 8.82 & 261.2 \\
OpenO1-Qwen & 9.53 & 265.9 \\
Open-Australian-Legal & 8.33 & 263.5 \\
DISC-LawLLM & 2.38 & 9.8 \\
Law-LLM & 4.18 & 97.4 \\
Law-Chat & 0.64 & 1.2 \\
Lawma & \textbf{0.63} & 1.0 \\
\cmidrule{1-3}
\textbf{Domain-CoT-finance} & 8.17 & 268.5 \\
\textbf{Domain$o1$s-legal} & 13.54 & \textbf{269.8} \\
\bottomrule
\end{tabular}
}
\caption{Inference time and reasoning chain length on legal reasoning QA tasks.}
\label{tab:time_length_legal}
\end{table}

\section{Answer Length and Inference Time}
\label{time_length}
In this section, we present the reasoning chain length and inference time of Domain$o1$s and baselines in generating answers for stock investment recommendations and legal reasoning QA tasks. The reasoning chain length is measured by the average number of words rather than tokens in the responses to ensure fair comparison across different models. As shown in Table~\ref{tab:time_length_stock} and Table~\ref{tab:time_length_legal}, o1-type models (OpenO1-Llama, OpenO1-Qwen, and our Domain$o1$s) have longer reasoning chains than other baselines, among which our Domain$o1$s and Domain-CoT have the longest reasoning chains. Although Domain$o1$s exhibits longer inference time compared to the baselines, this is attributed to its generation of longer and higher-quality reasoning chains and the search for optimal reasoning paths, ultimately leading to superior accuracy metrics.

\section{PROOF-Score Generation}
\label{Appendix_PROOF-Score_Generation}
\begin{figure}[htbp]
\centering
\includegraphics[width=0.49\textwidth]{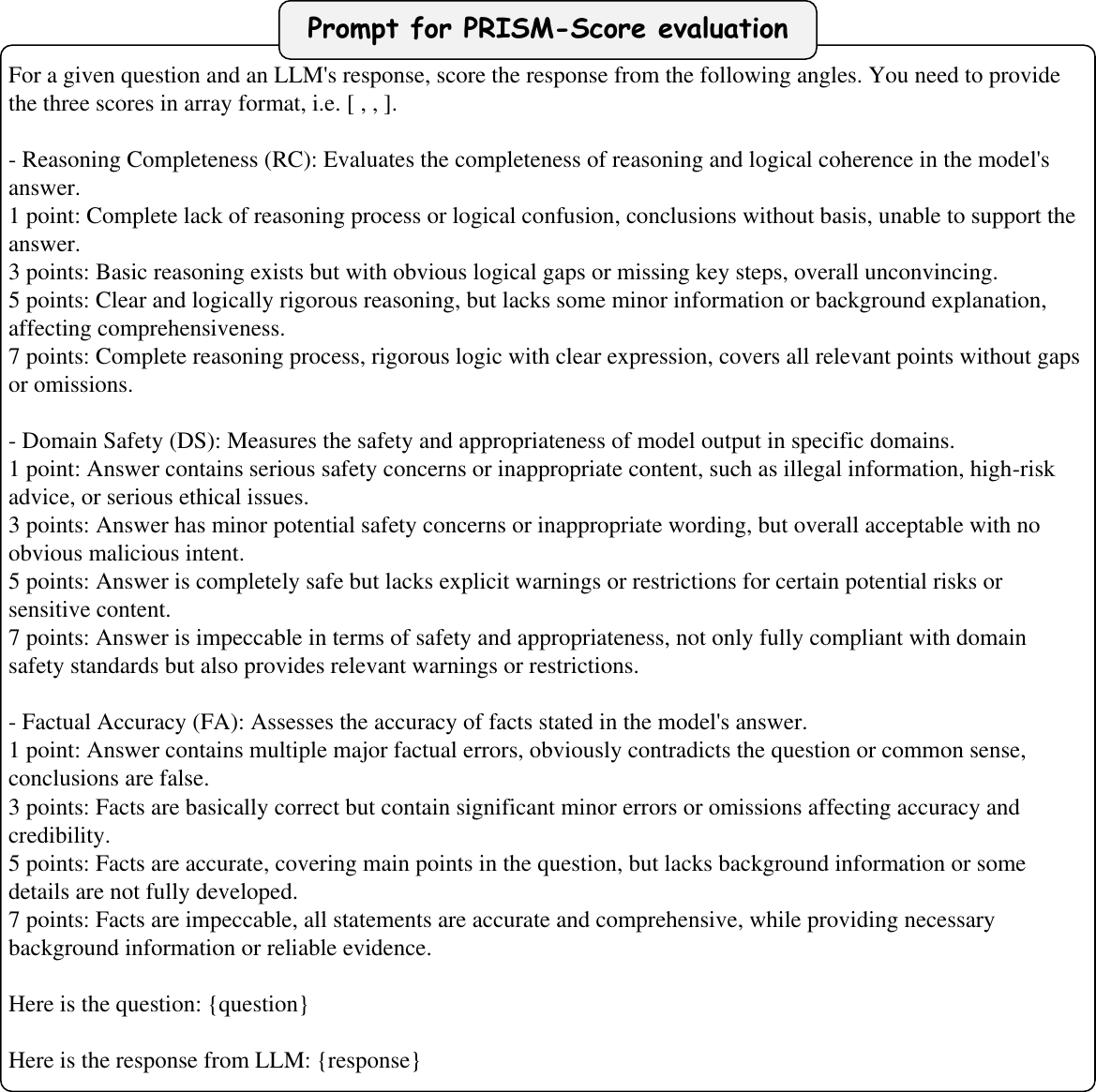}
\caption{Prompt template for generating PROOF-Score.}
\label{fig:prompt_prism}
\end{figure}

Figure~\ref{fig:prompt_prism} is the prompt template for instructing GPT-4o to generate PROOF-Scores.

\section{Construction of Sub-datasets}
\label{Appendix_Sub-datasets}
\begin{figure}[htbp]
\centering
\includegraphics[width=0.49\textwidth]{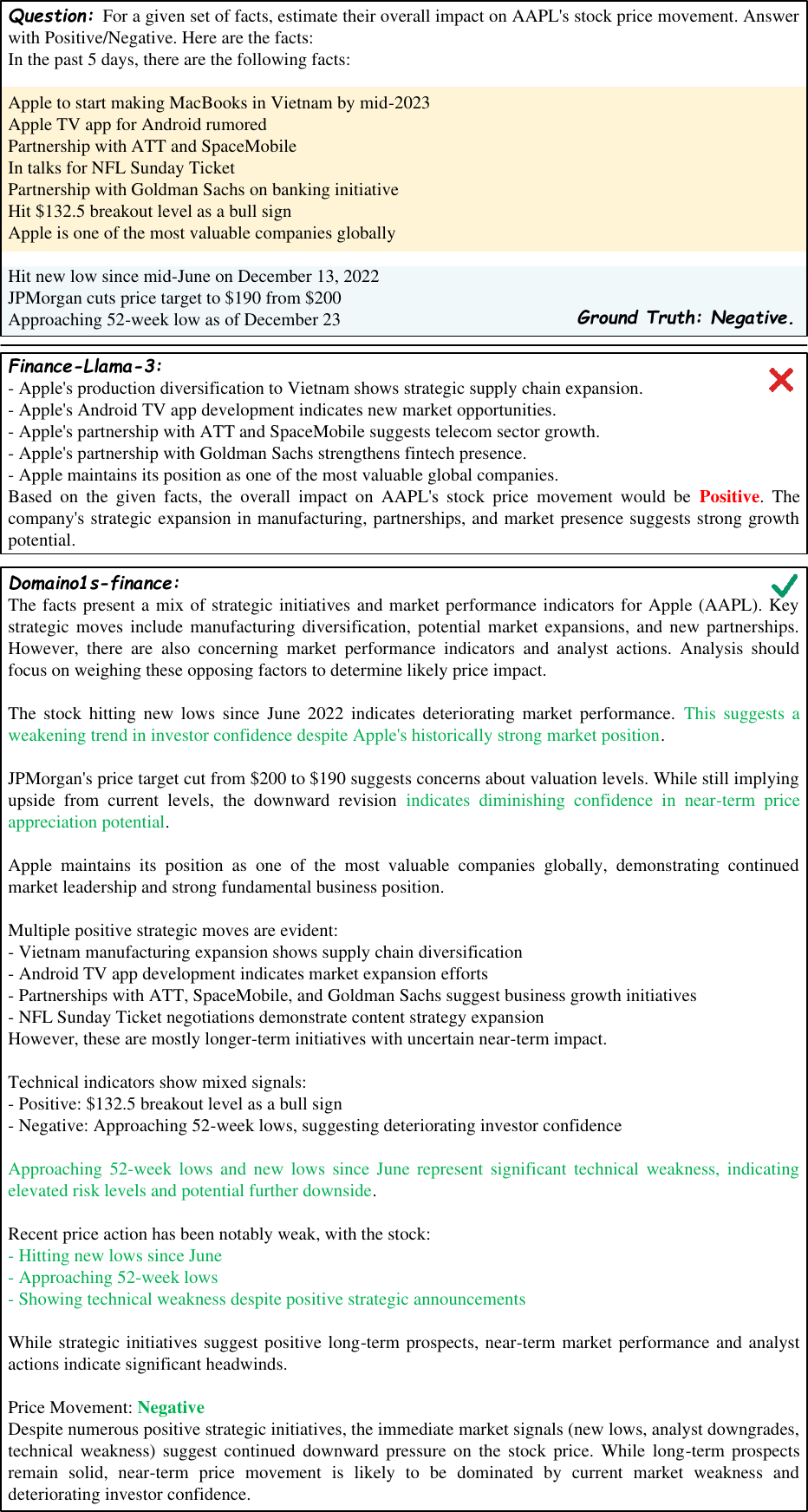}
\caption{Example of stock sub-dataset construction.}
\label{fig:subset_stock}
\end{figure}

We select 8 data points from each of the 55 stocks in the stock investment recommendation dataset. Each data point contains more than 10 positive tweets and 10 negative tweets. We construct five groups of tweets, with each group containing 10 tweets but different ratios of positive to negative tweets: 1:9 (1 positive, 9 negative), 3:7 (3 positive, 7 negative), 5:5 (5 positive, 5 negative), 7:3 (7 positive, 3 negative), and 9:1 (9 positive, 1 negative). Figure~\ref{fig:subset_stock} shows a construction example and model responses.
The example has a Positive:Negative ratio of 7:3. Tweets with yellow background are positive, while those with blue background are negative. Finance-Llama-3's response only considers the positive tweets, completely ignoring the negative ones. In contrast, Domain$o1$s-finance considers both positive and negative tweets to arrive at the correct answer.

\begin{figure}[htbp]
\centering
\includegraphics[width=0.49\textwidth]{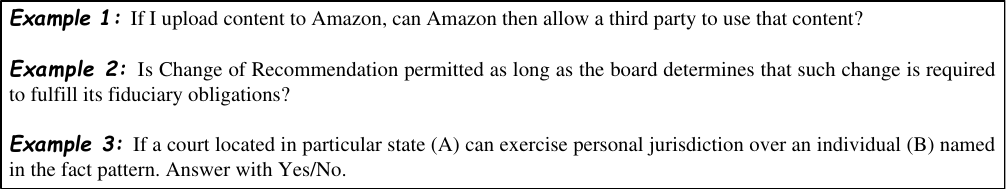}
\caption{Example of legal sub-dataset construction.}
\label{fig:subset_legal}
\end{figure}

We extract 500 questions from the legal reasoning QA dataset. Each question contains key conditions necessary for answering the question. We remove these key conditions from the questions, making them impossible to answer. Figure~\ref{fig:subset_legal} shows a construction example.

\section{Explainability Analysis}
\label{Appendix_Explainability}
Table~\ref{ablation_prism} demonstrates the ablation experiments of PROOF-Score without fine-tuning (w/o Fine-tune) and without solution expansion \& sampling (w/o Sample). In the w/o Fine-tune experiment, we use the Qwen-2.5-Instruct model without fine-tuning on our data and prompt it to separate each step with "\textbackslash n" to facilitate our solution expansion \& sampling. The results indicate that the PROOF-Score of the model without fine-tuning is lower than Domain$o1$s, demonstrating that Domain$o1$s learns to generate superior-quality reasoning processes from our constructed high-quality fine-tuning datasets. Meanwhile, the PROOF-Score of the model without solution expansion \& sampling is similar to Domain$o1$s, which suggests that the role of solution expansion \& sampling is more reflected in improving the quality of reasoning paths to enhance model accuracy (as shown in Table~\ref{tab_sample}-\ref{tab_theta}). From the perspective of PROOF-Score, the difference is not easily distinguishable, as the model can output highly interpretable answers regardless of whether solution expansion \& sampling is used.

\begin{table}[htbp]
\resizebox{\columnwidth}{!}{
\begin{tabular}{lc|cc}
\toprule
\multicolumn{2}{c}{\textbf{Stock}} & \multicolumn{2}{|c}{\textbf{Legal}} \\
\midrule
w/o Fine-tune & 6.212 & w/o Fine-tune & 5.067 \\
w/o Sample & 6.351 & w/o Sample & 6.548 \\
\cmidrule{1-4}
\textbf{Domain$o1$s-finance} & \textbf{6.359} & \textbf{Domain$o1$s-legal} & \textbf{6.677} \\
\bottomrule
\end{tabular}
}
\caption{Comparison of PROOF-Score between Domain$o1$s with w/o Fine-tune and w/o Sample.}
\label{ablation_prism}
\end{table}

\begin{table}[htbp]
\resizebox{\columnwidth}{!}{
\begin{tabular}{lcc|lcc}
\toprule
\multicolumn{3}{c|}{\textbf{Stock}} & \multicolumn{3}{c}{\textbf{Legal}} \\
\midrule
\textbf{Model} & \textbf{TIGERScore} & \textbf{Errors} & \textbf{Model} & \textbf{TIGERScore} & \textbf{Errors} \\
\midrule
Qwen-2.5-Instruct & 0.00 & 0.00 & Qwen-2.5-Instruct & -2.41 & 0.74 \\
Llama-3-Instruct & -0.50 & 0.50 & Llama-3-Instruct & -3.23 & 0.81 \\
OpenO1-Llama & 0.00 & 0.00 & OpenO1-Llama & -0.10 & 0.10 \\
OpenO1-Qwen & 0.00 & 0.00 & OpenO1-Qwen & -0.13 & 0.13 \\
Finance-LLM & -4.00 & 1.00 & Open-Australian-Legal & -6.40 & 1.60 \\
Finance-Chat & 0.00 & 0.00 & DISC-LawLLM & -4.00 & 1.00 \\
Finance-Llama-3 & -6.00 & 2.00 & Law-LLM & -2.45 & 1.11 \\
FinGPT & 0.00 & 0.00 & Law-Chat & -3.45 & 0.86 \\
Llama-2-taiwan-btc & 0.00 & 0.00 & Lawma & -3.76 & 0.94 \\
SEP & 0.00 & 0.00 & \textbf{Domain$o1$s-legal} & \textbf{-0.03} & \textbf{0.03} \\
\textbf{Domain$o1$s-finance} & \textbf{0.00} & \textbf{0.00} & & & \\
\bottomrule
\end{tabular}
}
\caption{Comparison of TIGERScore and error rates between Domain$o1$s and baselines on stock and legal tasks (using TIGERScore-7B). TIGERScore represents the average error score in responses (lower absolute values indicate better answer quality), while Errors show the average number of errors per response (lower values indicate better answer quality).}
\label{tab:tiger7b}
\end{table}

\begin{table}[htbp]
\resizebox{\columnwidth}{!}{
\begin{tabular}{lcc|lcc}
\toprule
\multicolumn{3}{c|}{\textbf{Stock}} & \multicolumn{3}{c}{\textbf{Legal}} \\
\midrule
\textbf{Model} & \textbf{TIGERScore} & \textbf{Errors} & \textbf{Model} & \textbf{TIGERScore} & \textbf{Errors} \\
\midrule
Qwen-2.5-Instruct & 0.00 & 0.00 & Qwen-2.5-Instruct & -0.76 & 0.19 \\
Llama-3-Instruct & 0.00 & 0.00 & Llama-3-Instruct & -1.26 & 0.32 \\
OpenO1-Llama & 0.00 & 0.00 & OpenO1-Llama & 0.00 & 0.00 \\
OpenO1-Qwen & 0.00 & 0.00 & OpenO1-Qwen & 0.00 & 0.00 \\
Finance-LLM & -4.00 & 1.00 & Open-Australian-Legal & -10.80 & 3.00 \\
Finance-Chat & -0.50 & 0.50 & DISC-LawLLM & -3.40 & 0.80 \\
Finance-Llama-3 & -8.00 & 2.00 & Law-LLM & -4.09 & 1.28 \\
FinGPT & 0.00 & 0.00 & Law-Chat & -1.10 & 0.27 \\
Llama-2-taiwan-btc & 0.00 & 0.00 & Lawma & -1.75 & 0.44 \\
SEP & 0.00 & 0.00 & \textbf{Domain$o1$s-legal} & \textbf{0.00} & \textbf{0.00} \\
\textbf{Domain$o1$s-finance} & \textbf{0.00} & \textbf{0.00} & & & \\
\bottomrule
\end{tabular}
}
\caption{Comparison of TIGERScore and error rates between Domain$o1$s and baselines on stock and legal tasks (using TIGERScore-13B). }
\label{tab:tiger13b}
\end{table}

In addition to our proposed PROOF-Score, we evaluate Domain$o1$s on other metrics. TIGERScore~\cite{jiang2023tigerscore} is an explainable reference-free evaluation metric based on LLaMA-2, which provides error analysis through natural language instructions and demonstrates the error analysis process. It can be used to evaluate a wide range of text-generation tasks. Table~\ref{tab:tiger7b} and Table~\ref{tab:tiger13b} show the evaluation results using TIGERScore-7B and TIGERScore-13B models respectively.

\begin{figure}[htbp]
\centering
\includegraphics[width=0.49\textwidth]{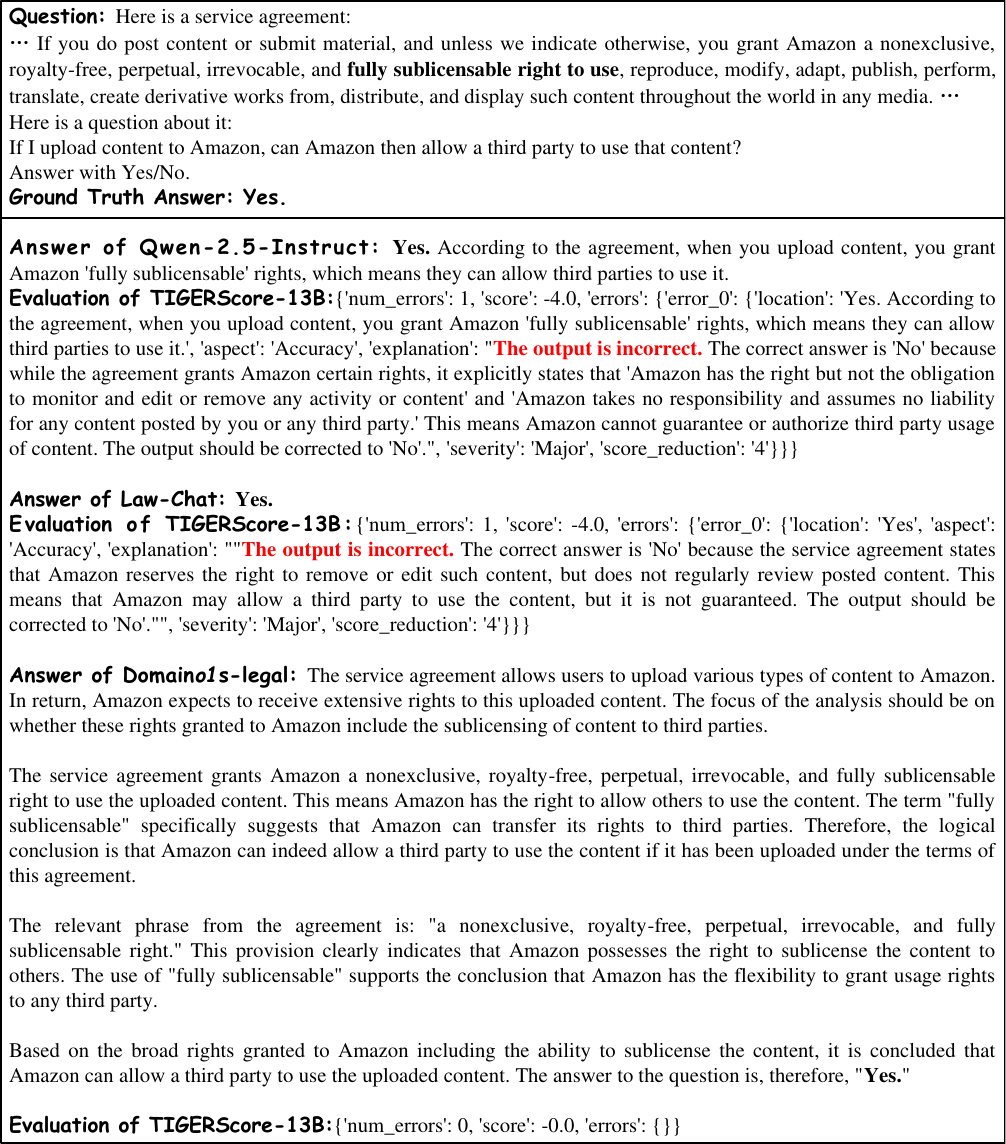} 
\caption{Example of TIGERScore-13B's evaluation.}
\label{tiger}
\end{figure}

As shown in Table~\ref{tab:tiger7b} and Table~\ref{tab:tiger13b}, Domain$o1$s achieves the highest scores (TIGERScore) and lowest error rates (Errors) in both stock and legal tasks, which indicates that Domain$o1$s' answers are more accurate and have more reasonable explanations compared to the baselines. Figure~\ref{tiger} shows examples of TIGERScore-13B's evaluation of answers from Domain$o1$s, Qwen-2.5-Instruct, and Law-Chat. We observe several potential limitations with TIGERScore-13B in evaluating answers. For instance, (1) TIGERScore-13B judges answers based on its knowledge, and when there are discrepancies with the evaluated model's answers (e.g., Qwen-2.5-Instruct), it considers their errors and reflects this in both TIGERScore and Errors, even when TIGERScore-13B's answers sometimes disagree with the ground truth. Additionally, (2) when the evaluated model (e.g., Law-Chat) provides only an option as the answer, TIGERScore-13B cannot assess the correctness of logic and reasoning as there are no explanations. In such cases, TIGERScore-13B can only score based on its judgment of answer correctness, which may lead to incorrect evaluations as mentioned in (1). In contrast, when the evaluated model (e.g., Domain$o1$s-legal) provides long comprehensive answers, TIGERScore-13B no longer strictly scores based on the model's conclusions. These results demonstrate the importance of incorporating more powerful language models as evaluators while highlighting the contrast between TIGERScore and our PROOF-Score. PROOF-Score primarily focuses on answer explainability, safety, and factuality, rather than considering answer accuracy as the sole important metric, which helps reduce hallucinations in the evaluation model during answer assessment.

\section{Reasoning Process Analysis}
\begin{figure}[ht]
\centering
\includegraphics[width=0.49\textwidth]{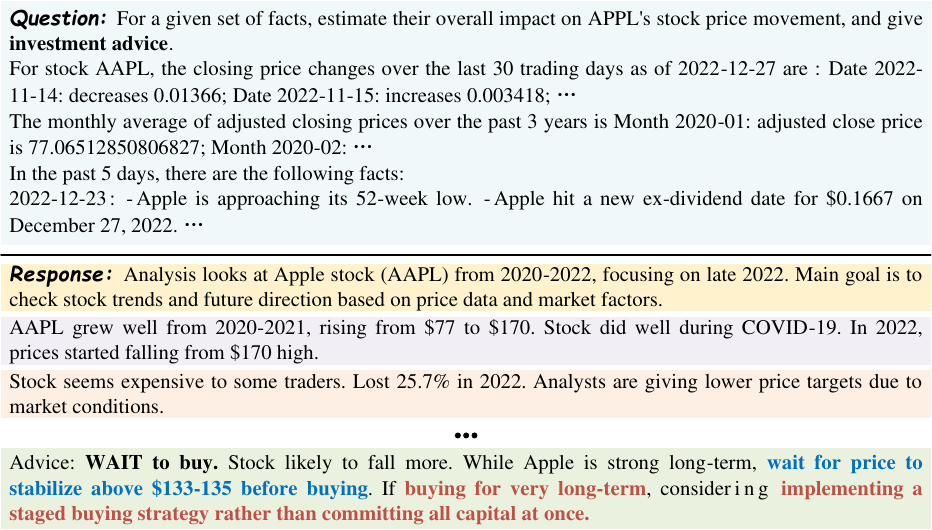} 
\caption{Example of Domain$o1$s-finance's response on stock investment recommendation. Blue bold text indicates short-term investment advice, and red bold text indicates long-term investment advice.}
\label{advice_exp}
\end{figure}

In this section, we introduce a new perspective to analyze how Domain$o1$s's reasoning process assists user decision-making. Taking stock investment recommendation tasks in the financial domain as an example, previous research has primarily focused on prediction accuracy as the main evaluation metric~\cite{koa2024learning,xu2018stock,sawhney2020deep,yang2022numhtml}. However, as shown in Table~\ref{tab_stock_main}, even the previously most advanced financial LLMs or LLMs extensively trained on stock data achieve prediction accuracy close to 50\% (random choice) for this binary classification problem. This confirms that LLMs are still "Wall Street Neophytes" in stock prediction~\cite{xie2023wall}. In reality, many domain problems in the real world often do not have unified or unique correct answers, as they typically involve dynamically changing environmental conditions, the interplay of multiple influencing factors, and user-specific preferences. When facing these problems, users can use Domain$o1$s's responses as references to aid their judgment by providing explainable analysis processes and recommendations. For example, Figure~\ref{advice_exp} shows part of the model's response when asking Domain$o1$s-finance for stock investment advice. The model's response includes both short-term and long-term analysis of stock prices and provides both short-term and long-term investment recommendations based on user investment habits.

\section{Training Data Scaling}
In this section, we explore the impact of increasing training data on Domain$o1$s performance. We introduce more legal QA data~\cite{li2022parameter}, and following the process in Section~\ref{sectionEnhancing Reasoning Capabilities}, we use GPT-4o to generate reasoning processes, expanding the CoT-legal dataset to 10k entries. Then, we train the Qwen-2.5-Instruction model with 2k, 5k, and 10k data separately, and evaluate the performance differences of Domain$o1$s-legal trained with different training data volumes. The results are as follows:

\begin{table}[h]
\resizebox{\columnwidth}{!}{
\begin{tabular}{lccccccccc>{\bfseries}c}
\toprule
\textbf{Model} & \textbf{CC} & \textbf{CAUD} & \textbf{MAUD} & \textbf{PP} & \textbf{IP} & \textbf{PJ} & \textbf{Scalar} & \textbf{TTP} & \textbf{TTD} & \textbf{Avg.} \\
\midrule
Domain$o1$s-legal (2k) & 88.64 & 81.76 & 80.33 & 66.54 & 52.63 & 72.00 & 88.97 & 95.33 & 78.78 & 78.33 \\
Domain$o1$s-legal (5k) & 88.73 & 81.74 & 80.65 & 66.95 & 51.88 & 74.00 & 89.32 & 93.93 & 77.57 & 78.30 \\
Domain$o1$s-legal (10k) & 88.96 & 81.63 & 81.87 & 66.63 & 53.38 & 74.00 & 89.32 & 93.93 & 80.37 & 78.90 \\
\bottomrule
\end{tabular}
}
\caption{Performance comparison of Domain$o1$s-legal models trained with different data volumes.}
\label{tab:data_scaling}
\end{table}

The results show that the accuracy improvement from increasing training data from 2k to 5k is not significant, and while increasing to 10k seems to increase the average accuracy (Avg.), the improvement on individual tasks is not substantial. This aligns with our approach of using a small number of training samples (2k) in this paper. The purpose of constructing a small number of training samples is not to introduce substantial domain-specific knowledge, but to stimulate the model's reasoning abilities in specific domains. However, we also encourage applying our method to domain models trained with large-scale domain data, using our few-shot approach to activate their reasoning capabilities to achieve stronger performance and robustness.

\section{Human Evaluation Alignment}
In Section~\ref{section:rq2}, we introduce PROOF-Scores based on GPT-4o to automate the evaluation of models' reasoning capability. Considering that human evaluations may have inconsistent scales and aspects compared to LLMs, we encourage having a human expert in the loop in evaluation scenarios that do not require rapid response and automation, to avoid biases introduced by LLMs.

On the other hand, studies~\cite{chiang2023can,zheng2023judging,desmond2024evalullm} support that LLMs like GPT-4 have evaluation capabilities comparable to humans. To understand whether this also applies to high-stake domain tasks, and to determine if our PROOF-Scores (GPT-4o) align with human judgment, we invited 3 PhD in finance and 3 in law to score the explainability of model results in Table~\ref{tab_reasoning} according to the scoring rules and intervals designed for PROOF-Scores. Given limited resources, we randomly sampled only 30\% of questions from each dataset to create the questionnaires.

For expert scoring, we calculate the PROOF-Score the same way as (\ref{eq:proofscore}). We calculate the average values and standard deviations of all expert scores, with results shown in Table~\ref{tab_reasoning2}.

\begin{table}[h]
\resizebox{\columnwidth}{!}{
\begin{tabular}{l|cc|l|cc}
\toprule
\multicolumn{3}{c|}{\textbf{Stock}} & \multicolumn{3}{c}{\textbf{Legal}} \\
\midrule
\textbf{Models} & \textbf{GPT-4o} & \textbf{Human expert} & \textbf{Models} & \textbf{GPT-4o} & \textbf{Human expert} \\
\midrule
Qwen-2.5-Instruct & 6.281 & 5.137±0.332 & Qwen-2.5-Instruct & 3.428 & 3.145±0.226 \\
Llama-3-Instruct & 6.129 & 5.025±0.248 & Llama-3-Instruct & 3.417 & 3.136±0.228 \\
OpenO1-Llama & 6.212 & 5.342±0.105 & OpenO1-Llama & 6.554 & 5.808±0.109 \\
OpenO1-Qwen & 6.227 & 5.358±0.133 & OpenO1-Qwen & 6.588 & 5.937±0.121 \\
Finance-LLM & 6.023 & 4.709±0.852 & Open-Australian-Legal & 5.152 & 2.382±0.536 \\
Finance-Chat & 5.583 & 4.832±0.648 & DISC-LawLLM & 0 & 1.403±0.310 \\
Finance-Llama3 & 5.965 & 4.813±0.731 & Law-LLM & 3.838 & 2.820±0.733 \\
FinGPT & 3.413 & 2.101±0.994 & Law-Chat & 3.339 & 3.274±0.384 \\
Llama-2-taiwan-btc & 0 & 1.438±0.455 & Lawma & 0 & 1.255±0.138 \\
SEP & 6.182 & 5.095±0.433 & & & \\
\midrule
\textbf{Domain$o1$s-finance} & \textbf{6.359} & \textbf{5.834±0.452} & \textbf{Domain$o1$s-Legal} & \textbf{6.677} & \textbf{6.296±0.148} \\
\bottomrule
\end{tabular}
}
\caption{Comparison of explanation quality (PROOF-Score) between Domain$o1$s and baselines as evaluated by GPT-4o and human experts.}
\label{tab_reasoning2}
\end{table}

The results show that although human expert scores are not completely consistent with GPT-4o, their relative magnitudes are similar. That is, the PROOF score (GPT-4o) assessment of the relative quality of different models is basically consistent with human experts. Additionally, both human experts and GPT-4o give Domain$o1$s the highest ratings.

\section{Application Extension}
We demonstrate that the method proposed in this paper suits finance and legal domains and can be extended to service recommendations, risk assessment, biological medicine, and other scenarios where explainable model answers are needed. Domain$o1$s can provide high-quality, interpretable answers for decision support in these fields.
\end{document}